%% file: main.tex
\documentclass{article}

\usepackage{iclr_conference,times}
\input{math_commands.tex}

\usepackage{latexsym}

\usepackage[T1]{fontenc}
\usepackage[utf8]{inputenc}

\usepackage{microtype}

\usepackage{graphicx}

\usepackage{enumitem}
\usepackage{booktabs}
\usepackage{adjustbox}
\usepackage{tabularray}
\usepackage{nicematrix}
\usepackage{multirow}
\usepackage{makecell}
\usepackage{threeparttable}
\usepackage{tabularx}
\usepackage{array}
\usepackage{amsmath}
\usepackage{amssymb}
\usepackage[table]{xcolor}
\usepackage{arydshln}
\usepackage{pifont}
\usepackage{algorithm}
\usepackage{algpseudocode}
\usepackage{float}
\usepackage{wrapfig}
\usepackage{placeins}
\usepackage{needspace}
\usepackage{mathtools}
\usepackage{fontawesome5}
\usepackage{subcaption}
\usepackage{hyperref}
\usepackage{url}
\hypersetup{
    colorlinks=true,
    citecolor=cyan,
    linkcolor=red,
    urlcolor=cyan
}

\newcommand{\AppendixRef}[1]{%
  \hyperref[#1]{\textcolor{black}{\textbf{\underline{Appendix~\textcolor{red}{\ref*{#1}}}}}}%
}
\newcommand{\AppendixTableRef}[1]{%
  \hyperref[#1]{\textcolor{black}{\textbf{\underline{Appendix Table~\textcolor{red}{\ref*{#1}}}}}}%
}
\newcommand{\AppendixAlgorithmRef}[1]{%
  \hyperref[#1]{\textcolor{black}{\textbf{\underline{Appendix Algorithm~\textcolor{red}{\ref*{#1}}}}}}%
}
\newcommand{\AppendixFigureRef}[1]{%
  \hyperref[#1]{\textcolor{black}{\textbf{\underline{Appendix Figure~\textcolor{red}{\ref*{#1}}}}}}%
}

\newcommand{\DefenseUp}[1]{%
  {\textcolor{red}{\scriptsize$\uparrow_{#1}$}}%
}
\newcommand{\DefenseDown}[1]{%
  {\textcolor{blue}{\scriptsize$\downarrow_{#1}$}}%
}

\newcommand{\DefenseAccUp}[1]{%
  {\textcolor{blue}{\scriptsize$\uparrow$#1}}%
}
\newcommand{\DefenseAccDown}[1]{%
  {\textcolor{red}{\scriptsize$\downarrow$#1}}%
}

\newcommand{\AblUp}[1]{\textcolor{red}{\scriptsize$\uparrow_{#1}$}}
\newcommand{\AblDown}[1]{\textcolor{blue}{\scriptsize$\downarrow_{#1}$}}

\definecolor{headercolor}{RGB}{211, 222, 219}

\definecolor{safecolor}{RGB}{204, 229, 255}
\definecolor{dangercolor}{RGB}{255, 204, 204}

\definecolor{safetextcolor}{RGB}{0, 82, 155}
\definecolor{dangertextcolor}{RGB}{155, 0, 0}

\newcommand{\safecell}[1]{\cellcolor{safecolor}\textcolor{safetextcolor}{\textbf{#1}}}
\newcommand{\dangercell}[1]{\cellcolor{dangercolor}\textcolor{dangertextcolor}{\textbf{#1}}}

\newcommand{\Best}[1]{\textbf{#1}}
\newcommand{\Second}[1]{\underline{#1}}
\newcommand{\secondbest}[1]{\underline{#1}}

\newcommand{\CSFRTuple}[4]{%
#1\,/\,#2\,/\,#3\,/\,#4%
}

\newcommand{\ObservationMark}{\ding{80}}
\newcommand{\TakeawayMark}{\ding{234}}

\title{Still There, No Longer Seen: Exposing Compression-Induced Risk in Large Vision-Language Models}

\iclrfinalcopy

\author{
\begin{tabular}{@{}l@{}}
\textbf{Qiankun Li\textsuperscript{1}}\thanks{Equal contribution.}
\quad
\textbf{Yuechen Zhang\textsuperscript{2}}\footnotemark[1]
\quad
\textbf{Bowen Chen\textsuperscript{2}}
\quad
\textbf{Shilinlu Yan\textsuperscript{2}}
\\
\textbf{Zhenhong Zhou\textsuperscript{1}}
\quad
\textbf{Kun Wang\textsuperscript{1}}\thanks{Kun Wang and Li Sun are the corresponding authors.}
\quad
\textbf{Li Sun\textsuperscript{2}}\footnotemark[2]
\end{tabular}
\\[2pt]
\textsuperscript{1}Nanyang Technological University
\\
\textsuperscript{2}Beijing University of Posts and Telecommunications
\\[2pt]
\texttt{\{cs-qiankun.li,wang.kun\}@ntu.edu.sg}
\quad
\texttt{orange.zhangyc05@gmail.com}
\\
\texttt{zhenhong001@e.ntu.edu.sg}
\quad
\texttt{\{cbcbw,lulu\_land,lsun\}@bupt.edu.cn}
}

\begin{document}
\maketitle

\input{1-abstract}
\input{2-introduction}
\input{3-related_work}
\input{4-motivation}
\input{5-method}

\input{6-experiments}

\input{7-defense}
\input{8-conclution}

\bibliographystyle{iclr_conference}
\bibliography{custom}

\newpage
\appendix
\input{9-appendix}

\end{document}

%% file: math_commands.tex
\usepackage{amsmath,amsfonts,bm}

\def\eqref#1{equation~\ref{#1}}
\def\Eqref#1{Equation~\ref{#1}}
\def\plaineqref#1{\ref{#1}}

\def\1{\bm{1}}

\DeclareMathAlphabet{\mathsfit}{\encodingdefault}{\sfdefault}{m}{sl}
\SetMathAlphabet{\mathsfit}{bold}{\encodingdefault}{\sfdefault}{bx}{n}

%% file: 1-abstract.tex
\begin{abstract}
Visual token compression reduces the inference cost of Large Vision-Language
Models (LVLMs). However, aggregate robustness measures do not reveal whether a
particular adversarial failure is induced by compression or inherited from the
underlying model. We define a \emph{compression-specific failure} (CSF) as an adversarial input
that remains correct under full-token inference but fails after compression,
casting compression-induced risk as a paired failure attribution problem. Within a controlled diagnostic cohort, counterfactuals show that retained-set
allocation causally changes compressed correctness and reveal a negative
association between recovery and representation drift in displaced evidence. Motivated by these findings, we propose
\textbf{CIRA}, a
\textbf{\underline{C}}ompression-\textbf{\underline{I}}nduced
\textbf{\underline{R}}isk \textbf{\underline{A}}ttack for Large Vision-Language
Models. Under a vision-encoder white-box setting, CIRA optimizes image perturbations
through encoder-side objectives that manipulate token priorities across
candidate compression budgets while preserving displaced evidence. CIRA uses no downstream questions or labels and requires no access to the
language model, deployed compressor, or exact compression budget. Across 12 dataset--compressor settings evaluated at four budgets, CIRA achieves
a mean CSFR of $20.35\%$ while limiting full-token attack success to $6.92\%$,
with similar behavior on additional LVLM families. A cross-view selection-stabilization defense substantially suppresses CIRA,
although Adaptive CIRA partially restores its effectiveness. These results show that compression-specific failures persist under restricted
access and support paired evaluation of full-token and compressed inference for
attributing risk to visual-token compression. Code is provided in the \href{https://github.com/RainNight11/CIRA}{\textcolor{magenta}{Github}}.
\end{abstract}

%% file: 2-introduction.tex
\section{Introduction}
\label{sec:introduction}

\begin{figure}[t]
  \centering
  \includegraphics[width=\textwidth]{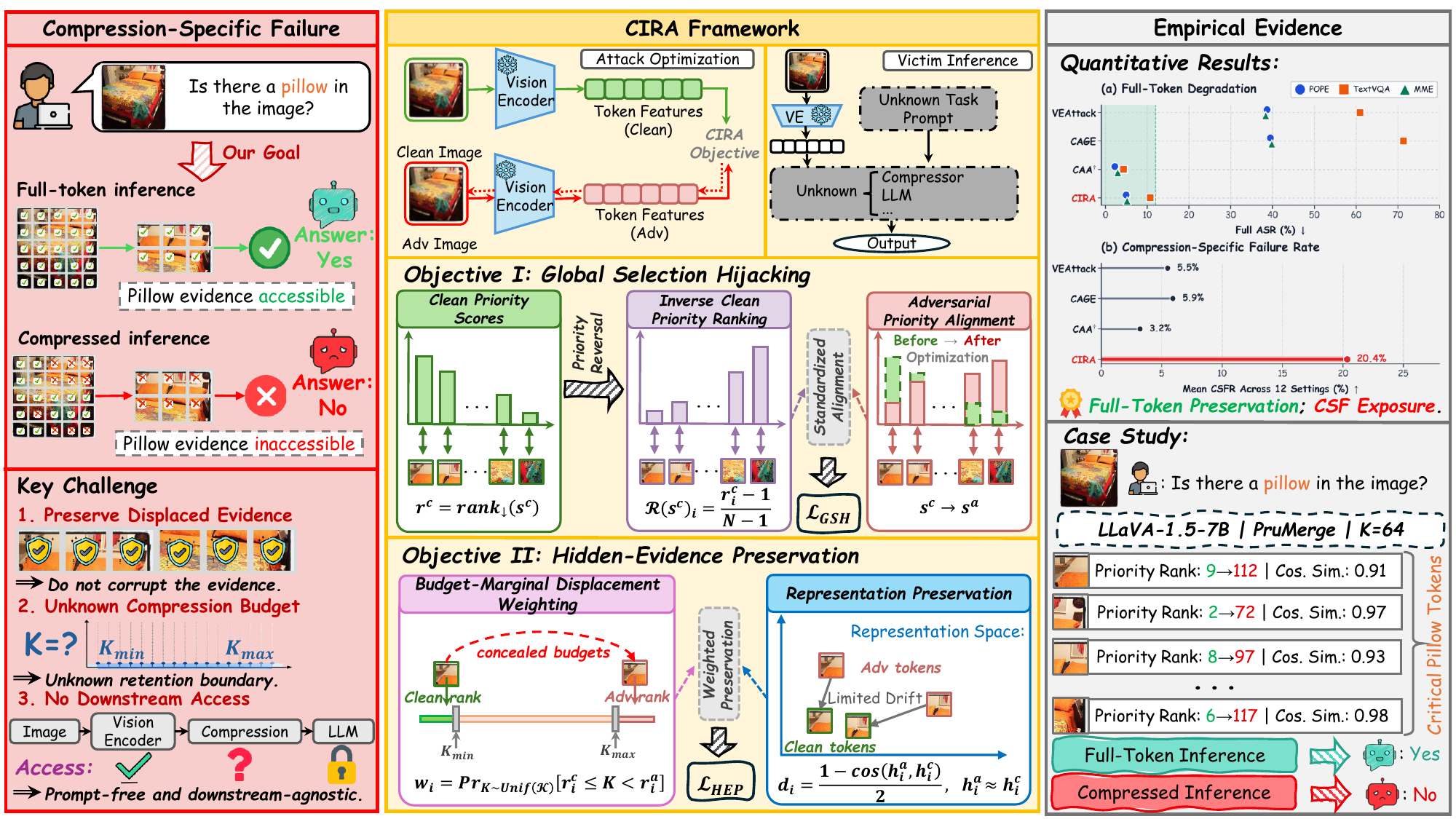}
  \caption{Overview of the compression-specific failure setting, the CIRA
  attack framework and the resulting behavior under visual-token compression.}
  \label{fig:framework}
\end{figure}

Large Vision-Language Models (LVLMs) support broad visual understanding by
mapping images to long visual-token sequences
\citep{alayrac2022flamingo,li2023blip,liu2024improved}. Visual-token
compression reduces this inference cost through token selection, aggregation,
and compact visual representations
\citep{shang2025llava,yang2025visionzip,li2025tokenpacker,bulat2026compress}. Decoder-side
methods further reduce visual-token computation within the language model
\citep{chen2024image,zhang2024sparsevlm,xing2024pyramiddrop}. Meanwhile,
adversarial attacks show that failures can be induced in the underlying
LVLM through image-space or joint-modal perturbations
\citep{zhang2022towards,schlarmann2023adversarial,yin2023vlattack}. This raises
the question:
\emph{what risks are induced by visual-token compression itself, beyond
vulnerabilities already present in the underlying LVLM?}

When a full-token LVLM is used for robustness assessment while a compressed
variant is deployed, failures confined to the compressed path are not
captured by that assessment. We therefore evaluate each adversarial input
with and without token compression. A \emph{compression-specific failure}
(CSF) occurs when full-token inference remains correct but compressed
inference fails. This paired criterion distinguishes failures introduced by
compression from those already present in the underlying LVLM.

Recent studies increasingly examine robustness under visual-token compression.
Evidence shows that compression can reshape robustness in either direction,
depending on how visual evidence is selected and retained
\citep{wang2026understanding,gu2026visual}. Compression-aware attacks further
demonstrate that adversarial behavior can depend on the compression path itself
\citep{zhang2026adversarial,zhang2026less}.
These developments shift the question from whether compression affects
robustness to which adversarial failures can be attributed specifically to
compression and whether they can be selectively induced under deployment
uncertainty. Specifically, can a single perturbation optimized through the target vision encoder preserve
full-token correctness while inducing compressed-path failures when the deployed
compressor and budget are unknown during optimization?

To understand what governs this selectivity, we use controlled
counterfactuals to isolate two factors. Retained-set interventions causally
change compressed correctness by altering which visual evidence remains
accessible after compression. Meanwhile, recovery is negatively associated
with representation drift in displaced evidence. These results directly inform
the attack design: reallocate token priority while preserving displaced
evidence.

In this direction, we introduce
\textbf{CIRA}
(\textbf{\underline{C}}ompression-\textbf{\underline{I}}nduced
\textbf{\underline{R}}isk \textbf{\underline{A}}ttack), an attack framework
for inducing compression-specific failures. As illustrated in
Figure~\ref{fig:framework}, CIRA combines Global Selection Hijacking (GSH),
which globally reallocates encoder-side proxy priorities, with Hidden-Evidence
Preservation (HEP), which limits representation drift in displaced clean
high-priority tokens. With white-box access restricted to the vision encoder,
CIRA uses no downstream questions or labels and requires no access to the
language model, deployed compressor, or exact compression budget.
Across four compressors, three datasets, four budgets, and multiple LVLMs,
CIRA induces compression-specific failures while keeping full-token degradation
limited. We also introduce Translation-Consensus Selection (TCS), a cross-view
selection-stabilization defense, and evaluate it against both standard and
adaptive CIRA.

Our contributions are threefold:
\begin{description}[labelwidth=0.35cm, leftmargin=!]

\item[\ding{182}] \textit{\underline{\textbf{Compression Risk Attribution.}}}
We formulate compression-induced risk through a clean-conditioned paired CSF
criterion and show that retained-set allocation causally affects compressed
correctness while recovery decreases with representation drift.

\item[\ding{183}] \textit{\underline{\textbf{Compression-Induced Risk Attack.}}}
We introduce CIRA, a target-encoder-only attack that combines priority
reallocation with evidence preservation and induces paired CSFs across
compressor--budget settings without configuration-specific optimization.

\item[\ding{184}] \textit{\underline{\textbf{Comprehensive Evaluation and Analysis.}}}
We evaluate CIRA across four compressors, three datasets, four budgets, and
multiple LVLM families, with mechanistic evidence of global priority
reallocation and limited representation drift in displaced evidence. We also
test TCS as a selection-stabilization defense against both standard and
adaptive CIRA.

\end{description}

%% file: 3-related_work.tex
\section{Related Work}
\label{sec:related_work}

\paragraph{Visual token compression.}
Token-selection methods exploit visual cues, diversity, and
salience--coverage objectives
\citep{zhang2025beyond,alvar2025divprune,xu2026score},
while learned pruning addresses limitations of attention-based importance
estimates \citep{takezoe2026learnpruner}.
Hybrid methods combine pruning with clustering and merging
\citep{endo2025feather,dhouib2025pact,yang2025libra}.
Query-conditioned methods use textual instructions for token scoring
or aggregation
\citep{yu2026visiontrim,gao2026quietprune,sun2026if}.
Adaptive and progressive methods vary token reduction across inputs and layers
\citep{ye2025atp,chen2026variation,li2026transprune},
while video and multi-turn approaches address temporal redundancy
and evolving context
\citep{wang2025dynamic,shen2024longvu,li2026vista,wang2026rethinking}.

\paragraph{Adversarial robustness of vision-language models.}
Transfer-based attacks exploit set-level guidance, prompt-robust objectives,
and iterative multimodal alignment
\citep{lu2023set,luo2024image,liu2024pandora,xie2025chain}.
Encoder-based and self-supervised attacks support cross-task and cross-model
transfer
\citep{zhang2025anyattack,hu2025transferable,zhang2026grounding}.
VEAttack disrupts visual representations \citep{mei2026veattack},
while PA-Attack uses prototype-guided gray-box attacks \citep{mei2026pa}.
Defenses include adversarial encoder fine-tuning \citep{schlarmann2024robust}
and adversarial pre-training and instruction tuning
\citep{wang2025double}.
Test-time defenses use prompt adaptation and augmented-view consistency
\citep{sheng2025r,liu2025self}.

\paragraph{Robustness under visual token compression.}
Visual-token compression introduces distinct robustness risks and
defense opportunities.
Safety-Aware Pruning (SAP) \citep{wang2026understanding} mitigates
pruning-induced vulnerabilities, while robustness-oriented pruning
\citep{gu2026visual} removes visually misaligned tokens.
On the attack side, CAGE \citep{zhang2026adversarial} targets tokens
expected to survive unknown compression settings.
CAA \citep{zhang2026less} directly manipulates token-selection rankings,
with white-box attacks tailored to known compression configurations
and transfer attacks using surrogate models.
Our focus is on selectively inducing compression-specific failures under
encoder-only access, without knowledge of the deployed compressor or budget.

%% file: 4-motivation.tex
\section{Diagnosing Compression-Specific Failures}
\label{sec:motivation}

In this section, we investigate why adversarial inputs can fail only after
compression and use controlled diagnostics to inform the design of CIRA.

\subsection{Problem Setup and Compression-Specific Failure}
\label{sec:compression_specific_failure}

Let $\mathcal{C}$ denote a visual-token compressor and $K$ its compression
budget. For a dataset
$\mathcal{D}=\{(x_i,q_i,y_i)\}_{i=1}^{M}$, let
$f_{\theta}(x_i,q_i)$ and
$f_{\theta}^{\mathcal{C},K}(x_i,q_i)$ denote full-token and compressed
inference, respectively. Given a task-specific binary evaluator
$\chi(\hat{y},y)\in\{0,1\}$, we define
\[
c_i(x)
=
\chi\!\left(f_{\theta}(x,q_i),y_i\right),
\qquad
c_{i,\mathcal{C},K}(x)
=
\chi\!\left(f_{\theta}^{\mathcal{C},K}(x,q_i),y_i\right).
\]
For a fixed $\mathcal{C}$, we write
$c_{i,K}(x)\equiv c_{i,\mathcal{C},K}(x)$ and omit the compressor subscript
below.

For each compression budget $K$, we restrict evaluation to samples for which both full-token and compressed inference produce correct
answers before attack:
\begin{equation}
\mathcal{S}_{K}
=
\left\{
i
\;\middle|\;
c_i(x_i)=1,\;
c_{i,K}(x_i)=1
\right\}.
\label{eq:clean_eligible_set}
\end{equation}

A \emph{compression-specific failure} (CSF) occurs when an adversarial input
remains correct under full-token inference but fails after compression:
\begin{equation}
\mathcal{F}^{\mathrm{CSF}}_{K}
=
\left\{
i\in\mathcal{S}_{K}
\;\middle|\;
c_i(x_i^{\mathrm{adv}})=1,\;
c_{i,K}(x_i^{\mathrm{adv}})=0
\right\}.
\label{eq:compression_specific_failure}
\end{equation}

Aggregate accuracy can obscure instance-level prediction changes
introduced by VLM acceleration \citep{sun2025does}.
Our paired definition focuses on adversarial inputs that remain correct
under full-token inference but fail after compression.

\subsection{Retained-Set Allocation and Compressed Correctness}
\label{sec:selection_surface}

\paragraph{Diagnostic setting.}
We test whether retained-set allocation can change compressed correctness while
the adversarial encoder state and compression capacity remain fixed.
LLaVA--VisionZip exposes direct-token membership, providing a controlled
interface for retained-set intervention. From each downstream-agnostic VEAttack trajectory, we retain the earliest checkpoint satisfying the CSF criterion, termed a \emph{first-observed CSF}.

\paragraph{Counterfactual intervention.}
For each CSF, we exchange equal numbers of retained and omitted direct tokens,
reconstruct compression, and rerun inference at the same encoder state and
capacity. A post-hoc answer-aware oracle selects the exchanged token identities
(\AppendixRef{sec:appendix_motivation}).

\paragraph{Recovery criteria.}
Let $\rho$ denote the fraction of exchanged direct-token slots. We compare
guided reallocation with a size-matched random exchange. Cumulative recovery
credits correction at any tested $\rho'\leq\rho$, whereas exact recovery uses
only $\rho$.

\paragraph{\ObservationMark\ \textbf{Observation 1:} Controlled retained-set reallocation changes compressed correctness.}

\Needspace{22\baselineskip}
\begin{wrapfigure}[15]{r}{0.49\textwidth}
    \vspace{-12pt}
    \centering
    \includegraphics[width=\linewidth]{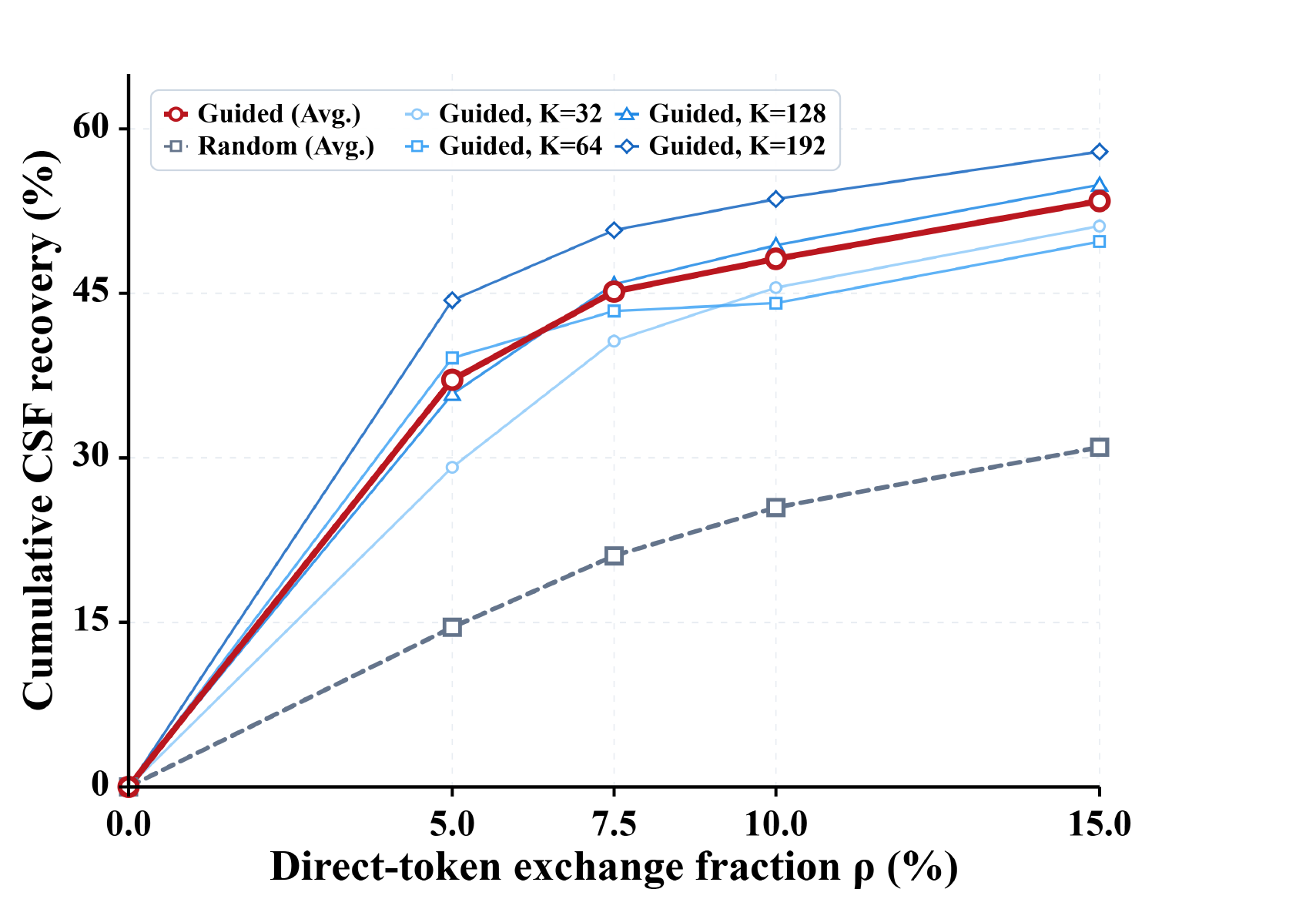}
    \caption{
    Cumulative CSF recovery under guided and matched-random retained-set exchanges
    across direct-token exchange fractions.
    }
    \label{fig:repairability}
\end{wrapfigure}

Figure~\ref{fig:repairability} shows that guided recovery exceeds matched-random recovery
for every tested budget--fraction pair with $\rho>0$, with an average advantage of
$22.5$--$24.1$ pp across the 12 dataset--budget cells. With encoder state,
capacity, and exchange size fixed, the contrast shows that robustness depends
on token identity, not only on retention count.

\subsection{Retained-Set Allocation Components and Representation Drift}
\label{sec:selection_effects}

Having established retained-set allocation sensitivity, we next separate
restoration from removal and ask when restored evidence remains useful.
Observations 2--3 use $\rho^{*}=7.5\%$ as a shared intermediate diagnostic
point, where guided and random recovery are already clearly separated.
Representation drift is the restored tokens' mean half-cosine distance,
$[1-\cos(\mathbf h^c,\mathbf h^a)]/2$, between their clean and adversarial
representations; Table~\ref{tab:evidence_source_analysis}(b) reports this
distance multiplied by $100$.

\paragraph{\ObservationMark\ \textbf{Observation 2: Recovery depends on what is restored and removed.}}
Table~\ref{tab:evidence_source_analysis}(a) decomposes guided reallocation
into restoring omitted tokens with high retrospective answer support and
removing retained tokens with low support; larger scores indicate stronger
support for the reference answer. Exact recovery rises from $17.4\%$ under
matched random exchange to $42.0\%$ under guided reallocation. Both
restoration and removal contribute to recovery, with their coordinated use
providing the largest gain.

\input{tables/mechanistic_diagnostics}

\paragraph{\ObservationMark\ \textbf{Observation 3:} Restoration benefits are smaller under greater representation drift.}
Table~\ref{tab:evidence_source_analysis}(b) shows that evidence restoration is
most effective for tokens with limited representation drift: the average effect
falls from $+20.5$ pp in the lower-drift tertile to $+10.3$ pp in the upper.
A within-cell continuous analysis shows the same negative association.

\paragraph{Design implication.}
Together, these diagnostics motivate CIRA's objective: make evidence
inaccessible after compression while preserving its utility under full-token
inference.

%% file: tables/mechanistic_diagnostics.tex
\begin{table*}[htbp]
\centering
\scriptsize
\vspace{-3pt}

\caption{
Mechanistic diagnostics at $\rho^{*}=7.5\%$:
exact recovery under retained-set interventions (a) and evidence-restoration
effects across representation-drift tertiles (b), with construction and
aggregation details in \AppendixRef{sec:appendix_motivation}.
}
\label{tab:evidence_source_analysis}

\vspace{1pt}

\begin{minipage}[t]{0.49\textwidth}
\centering
{\footnotesize\bfseries (a) Retained-Set Counterfactuals}
\end{minipage}
\hfill
\begin{minipage}[t]{0.49\textwidth}
\centering
{\footnotesize\bfseries (b) Association with Representation Drift}
\end{minipage}

\vspace{0pt}

\begin{adjustbox}{max width=\textwidth}

\begin{tabular}{
@{}
c
@{\hspace{0.55cm}}
c
@{}
}

% Left table: counterfactual composition
{
\renewcommand{\arraystretch}{1.16}
\setlength{\tabcolsep}{3.0pt}

\begin{tabular}{
>{\raggedright\arraybackslash}m{2.65cm}
!{\color{black}\vrule width 0.45pt}
cccc
!{\color{black}\vrule width 0.45pt}
c
}

\Xhline{2pt}

\rowcolor{headercolor}
\textbf{Retained-Set Change}
&
\multicolumn{4}{
c!{\color{black}\vrule width 0.45pt}
}{
\textbf{Exact Recovery (\%)}
}
&
\textbf{Avg.}
\\

\rowcolor{headercolor}
{}
& $K=192$
& $K=128$
& $K=64$
& $K=32$
& {}
\\

\Xhline{2pt}

Random
& 17.9
& 17.2
& 18.4
& 16.1
& \textbf{17.4}
\\

\hspace{0.45em}$+$ Evidence restoration
& 31.1
& 33.5
& 34.0
& 34.9
& \textbf{33.4}
\\

\hspace{0.45em}$+$ Low-support removal
& 34.5
& 31.9
& 25.4
& 20.1
& \textbf{28.0}
\\

\Xhline{0.45pt}

\rowcolor{gray!12}
\textbf{Guided reallocation}
& 46.4
& 44.4
& 38.5
& 38.6
& \textbf{42.0}
\\

\Xhline{2pt}

\end{tabular}
}

&

% Right table: representation drift
{
\renewcommand{\arraystretch}{1.16}
\setlength{\tabcolsep}{3.0pt}

\begin{tabular}{
c
!{\color{black}\vrule width 0.45pt}
c
!{\color{black}\vrule width 0.45pt}
cccc
!{\color{black}\vrule width 0.45pt}
c
}

\Xhline{2pt}

\rowcolor{headercolor}
\textbf{Drift Tertile}
&
\textbf{Mean Drift (\%)}
&
\multicolumn{4}{
c!{\color{black}\vrule width 0.45pt}
}{
\textbf{Evidence Effect (pp)}
}
&
\textbf{Avg.}
\\

\rowcolor{headercolor}
{}
& {}
& $K=192$
& $K=128$
& $K=64$
& $K=32$
& {}
\\

\Xhline{2pt}

Lower
& 31.4
& $+16.6$
& $+18.9$
& $+18.2$
& $+28.5$
& $\mathbf{+20.5}$
\\

Middle
& 39.5
& $+15.3$
& $+12.7$
& $+7.0$
& $+20.8$
& $\mathbf{+14.0}$
\\

Upper
& 45.2
& $+5.4$
& $+11.5$
& $+18.0$
& $+6.1$
& $\mathbf{+10.3}$
\\

\Xhline{0.45pt}

\rowcolor{gray!12}
\textbf{Lower--Upper Gap}
& --
& $+11.2$
& $+7.4$
& $+0.2$
& $+22.3$
& $\mathbf{+10.3}$
\\

\Xhline{2pt}

\end{tabular}
}

\end{tabular}

\end{adjustbox}
\vspace{-3pt}

\end{table*}

%% file: 5-method.tex
\section{Methodology}
\label{sec:methodology}

\subsection{Attack Formulation}
\label{sec:attack_formulation}

We consider an untargeted, image-specific attack with white-box access only to
the vision encoder. The downstream question, reference answer, language model,
compressor, and exact deployed compression budget are unavailable during
optimization; the attacker knows only the admissible range
$\mathcal{K}=\{K\in\mathbb{N}:K_{\min}\leq K\leq K_{\max}\}$.

For an evaluation tuple $(x,q,y)$, the ideal compression-specific outcome is
to preserve full-token correctness while causing compressed inference to fail:
\begin{equation}
\max_{\|\delta\|_{\infty}\leq\epsilon}
\;
\min_{K\in\mathcal{K}}
\mathcal{L}
\left(
f_{\theta}^{\mathcal{C},K}(x+\delta,q),y
\right)
\quad
\mathrm{s.t.}\quad
c(x+\delta)=1,
\label{eq:attack_objective}
\end{equation}
where $c(\cdot)$ denotes full-token correctness and $\mathcal{L}$ increases with
compressed prediction error. \Eqref{eq:attack_objective} defines the desired compression-specific
behavior but is not optimized directly. CIRA instead replaces it with
compressor-independent encoder-side objectives and evaluates cross-configuration
transfer empirically.

\subsection{Global Selection Hijacking}
\label{sec:selection_hijacking}

Let $s_i(x)$ denote the encoder-side proxy priority score of visual token $i$,
where larger values indicate higher retention priority. We collect these scores
in the priority-score vector
$\mathbf{s}(x)=[s_1(x),\ldots,s_N(x)]\in\mathbb{R}^{N}$, whose descending
order defines the priority ranking over tokens. We write
$\mathbf{s}^{c}=\mathbf{s}(x)$ and
$\mathbf{s}^{a}=\mathbf{s}(x+\delta)$
for the clean and adversarial priority-score vectors, respectively.

We first construct the clean rank, its inverse-priority encoding, and a
stabilized standardization operator:
\begin{equation}
\mathbf{r}^{c}
=
\operatorname{rank}_{\downarrow}(\mathbf{s}^{c}),
\qquad
\mathcal{R}(\mathbf{s}^{c})_i
=
\frac{r_i^c-1}{N-1},
\qquad
\mathcal{Z}_{\epsilon_s}(\mathbf{u})
=
\frac{\mathbf{u}-\bar{u}\mathbf{1}}
{\max\!\left\{
N^{-1/2}\|\mathbf{u}-\bar{u}\mathbf{1}\|_2,\epsilon_s
\right\}} .
\label{eq:gsh_transform}
\end{equation}
Here, $\mathbf{r}^{c}$ ranks tokens from high to low clean priority, while
$\mathcal{R}$ maps high-priority tokens near zero and low-priority tokens near
one. $\mathcal{Z}_{\epsilon_s}$ standardizes its input vector and prevents a
degenerate denominator when its variance vanishes.

To drive a global priority inversion, \textsc{CIRA} maximizes the standardized
alignment between adversarial priorities and the inverse clean priority ranking:
\begin{equation}
\mathcal{L}_{\mathrm{GSH}}(\delta)
=
\operatorname{Align}_{\epsilon_s}
\!\left(
\mathbf{s}^{a},
\mathcal{R}(\mathbf{s}^{c})
\right)
\equiv
\frac{1}{N}
\left\langle
\mathcal{Z}_{\epsilon_s}(\mathbf{s}^{a}),
\mathcal{Z}_{\epsilon_s}\!\left(\mathcal{R}(\mathbf{s}^{c})\right)
\right\rangle .
\label{eq:gsh_loss}
\end{equation}
This differentiable, scale-invariant objective encourages clean high-priority
tokens to move downward while promoting clean low-priority tokens.

\subsection{Hidden-Evidence Preservation}
\label{sec:hidden_evidence}

Priority reallocation may also perturb the representations of displaced
evidence, undermining full-token correctness. HEP therefore focuses
preservation on clean high-priority tokens that cross candidate retention
boundaries. \textsc{CIRA} computes the adversarial descending ranks
$\mathbf{r}^{a}=\operatorname{rank}_{\downarrow}(\mathbf{s}^{a})$ and defines
$\mathcal{H}_{K}(\delta)=\{i:r_i^c\leq K<r_i^a\}$, the clean proxy Top-$K$
tokens whose adversarial ranks move beyond the retention boundary at
compression budget $K$. Under an unknown compression budget, their displacement
weights are
\begin{equation}
w_i
=
\Pr_{K\sim\mathrm{Unif}(\mathcal{K})}
\!\left[i\in\mathcal{H}_{K}(\delta)\right]
=
\frac{
\left[
\min(K_{\max},r_i^a-1)
-\max(K_{\min},r_i^c)+1
\right]_{+}
}{
K_{\max}-K_{\min}+1
}.
\label{eq:hidden_weight}
\end{equation}
Here, $[z]_{+}=\max(z,0)$, and $w_i$ is the fraction of admissible budgets
under which token $i$ becomes hidden.

To limit representation drift in these tokens, let $\mathbf{h}_i^c$ and
$\mathbf{h}_i^a$ denote their clean and adversarial visual-token features.
We normalize their directions and
measure the resulting representation drift by
\begin{equation}
\widetilde{\mathbf{h}}_i^{\,c}
=
\frac{\mathbf{h}_i^c}{\|\mathbf{h}_i^c\|_2},
\qquad
\widetilde{\mathbf{h}}_i^{\,a}
=
\frac{\mathbf{h}_i^a}{\|\mathbf{h}_i^a\|_2},
\qquad
d_i
=
\frac{1}{2}
\left(
1-
(\widetilde{\mathbf{h}}_i^{\,c})^{\top}
\widetilde{\mathbf{h}}_i^{\,a}
\right),
\label{eq:hidden_drift}
\end{equation}
where the factor $1/2$ normalizes cosine distance to $[0,1]$.

Hidden-Evidence Preservation aggregates these token-level distances using the
budget-marginal weights:
\begin{equation}
\mathcal{L}_{\mathrm{HEP}}(\delta)
=
-\sum_{i=1}^{N}\widetilde{w}_i d_i,
\qquad
\widetilde{w}_i
=
\frac{
\operatorname{sg}(w_i)
}{
\max\!\left\{
\sum_{j=1}^{N}\operatorname{sg}(w_j),\epsilon_h
\right\}
}.
\label{eq:hep_loss}
\end{equation}
The stop-gradient freezes rank-derived weights within an update, while
$\epsilon_h$ stabilizes normalization; weights are recomputed at the next update
so evidence displaced across more of $\mathcal{K}$ receives greater protection.

\subsection{Joint Optimization}
\label{sec:joint_optimization}

\textsc{CIRA} combines the two objectives as
\begin{equation}
\max_{\delta}
\quad
\mathcal{L}_{\mathrm{CIRA}}(\delta)
=
\mathcal{L}_{\mathrm{GSH}}(\delta)
+
\lambda \cdot \mathcal{L}_{\mathrm{HEP}}(\delta),
\qquad
\mathrm{s.t.}\quad
\|\delta\|_{\infty}\leq\epsilon .
\label{eq:joint_cira_loss}
\end{equation}
Here, $\lambda$ controls the preservation strength. We maximize \eqref{eq:joint_cira_loss} using projected sign-gradient ascent
over the valid-image domain. The complete optimization procedure is given in
\AppendixRef{sec:appendix_cira}.

%% file: 6-experiments.tex
\section{Experiments}
\label{sec:experiments}

\subsection{Experimental Setup}
\label{sec:experimental_settings}

\paragraph{Models and Benchmarks.}
We evaluate LLaVA-v1.5-7B \citep{liu2024improved} on 1,000 randomly sampled
image--question pairs from each of POPE \citep{li2023evaluating}, TextVQA
\citep{singh2019towards}, and MME \citep{fu2026mme}, covering object
hallucination, scene-text understanding, and general visual perception and
reasoning, respectively. We additionally evaluate Qwen3-VL-8B-Instruct
\citep{bai2025qwen3} and InternVL3.5-8B \citep{wang2025internvl3}.

\paragraph{Compression Settings.}
We evaluate VisionZip \citep{yang2025visionzip}, VisPruner
\citep{zhang2025beyond}, PruMerge \citep{shang2025llava}, and FastV
\citep{chen2024image} at
$\mathcal{K}_{\mathrm{eval}}=\{32,64,128,192\}$.
For each input, one adversarial image is reused across all compressor--budget
settings.

\paragraph{Baselines.}
We compare CIRA with the downstream-agnostic VEAttack
\citep{mei2026veattack} and CAGE \citep{zhang2026adversarial}; CAA
\citep{zhang2026less} is reported as a stronger-access reference.

\paragraph{Evaluation Metrics.}
\label{sec:evaluation_metrics}
On the clean-eligible set $\mathcal{S}_{K}$ defined in
Eq.~\plaineqref{eq:clean_eligible_set}, we report the Compression-Specific
Failure Rate (CSFR) and full-token attack success rate:
\begin{equation}
\mathrm{CSFR}_{K}
=
\frac{1}{|\mathcal{S}_{K}|}
\sum_{i\in\mathcal{S}_{K}}
c_i(x_i^{\mathrm{adv}})
\left[
1-c_{i,K}(x_i^{\mathrm{adv}})
\right],
\qquad
\mathrm{ASR}
=
\frac{1}{|\mathcal{S}_{\mathrm{full}}|}
\sum_{i\in\mathcal{S}_{\mathrm{full}}}
\left[
1-c_i(x_i^{\mathrm{adv}})
\right].
\label{eq:evaluation_metrics}
\end{equation}
Here, $\mathcal{S}_{\mathrm{full}}=\{i:c_i(x_i)=1\}$, and Avg.\ CSFR is the
arithmetic mean of $\mathrm{CSFR}_{K}$ over
$K\in\mathcal{K}_{\mathrm{eval}}$ for a fixed dataset--compressor setting.
Broader summaries weight each reported dataset--compressor setting equally.
CSFR counts post-attack failures confined to the compressed path, normalized
over inputs answered correctly by both clean inference paths; ASR is normalized
over inputs answered correctly by clean full-token inference. Clean and post-attack accuracy results are reported in
\AppendixRef{sec:appendix_task_utility}.

\paragraph{Implementation Settings.}
We run all experiments on a single NVIDIA GeForce RTX 4090 GPU.
All downstream-agnostic attacks use $\epsilon=4/255$ and 100 optimization
steps; CIRA uses projected sign-gradient ascent with $\lambda=0.8$.
Further protocol details and sensitivity analyses appear in
\AppendixRef{sec:appendix_detailed_setup} and
\AppendixRef{sec:appendix_sensitivity}.

\subsection{Main Results}
\label{sec:main_results}

In this section, we evaluate CIRA's compression selectivity and
cross-configuration transfer, and compare it with stronger-access attacks.

\input{tables/main_results}

\paragraph{\textbf{Compression-specific selectivity.}}
Across the four budgets and 12 dataset--compressor settings in
Table~\ref{tab:results_csfr}, CIRA achieves a mean CSFR of $20.35\%$, compared
with $5.49\%$ for VEAttack and $5.92\%$ for CAGE. Its Full ASR is $6.92\%$,
substantially below $45.95\%$ and $50.20\%$, respectively. CIRA therefore induces more compression-specific failures while causing much less full-token degradation than either downstream-agnostic baseline.

\paragraph{\textbf{Cross-configuration transfer.}}
The same adversarial image is reused without re-optimization across
selection-, pruning-, and merging-based compression rules and all evaluated
budgets. CIRA remains effective across these configurations, reaching
$31.96\%$ mean CSFR on PruMerge compared with $6.28\%$ for CAGE. Averaged
equally over all 12 dataset--compressor settings, its CSFR increases from
$12.04\%$ at $K=192$ to $28.78\%$ at $K=32$. Compression-selective behavior
also extends to Qwen3-VL and InternVL
(Table~\ref{tab:qwen_internvl_compact_results}); qualitative cross-setting
examples appear in \AppendixRef{sec:case-studies}.

\paragraph{\textbf{Comparison with stronger-access CAA.}}
CAA$^{\dagger}$ optimizes with access to the downstream question and language
model, whereas CIRA uses only the vision encoder. Despite its lower Full ASR
($3.32\%$ for CAA$^{\dagger}$ versus $6.92\%$ for CIRA), CAA$^{\dagger}$
achieves only $3.24\%$ mean CSFR, well below CIRA's $20.35\%$.
The contrast shows that preserving the full-token prediction is not sufficient
to induce compression-specific failures; CIRA's priority-reallocation
objective addresses this distinct requirement under narrower access.

\paragraph{\TakeawayMark\ \textbf{Takeaway.}}
CIRA induces compression-specific failures across heterogeneous compression
rules and budgets while limiting full-token degradation.

\subsection{Mechanistic Analysis}
\label{sec:mechanism_analysis_exp}

To connect CIRA's behavior to its objectives, we test whether it globally
reallocates token priority while limiting representation drift in displaced
evidence.

\begin{figure*}[t]
    \centering

    \begin{minipage}[t]{0.58\textwidth}
        \centering
        \includegraphics[width=\linewidth]{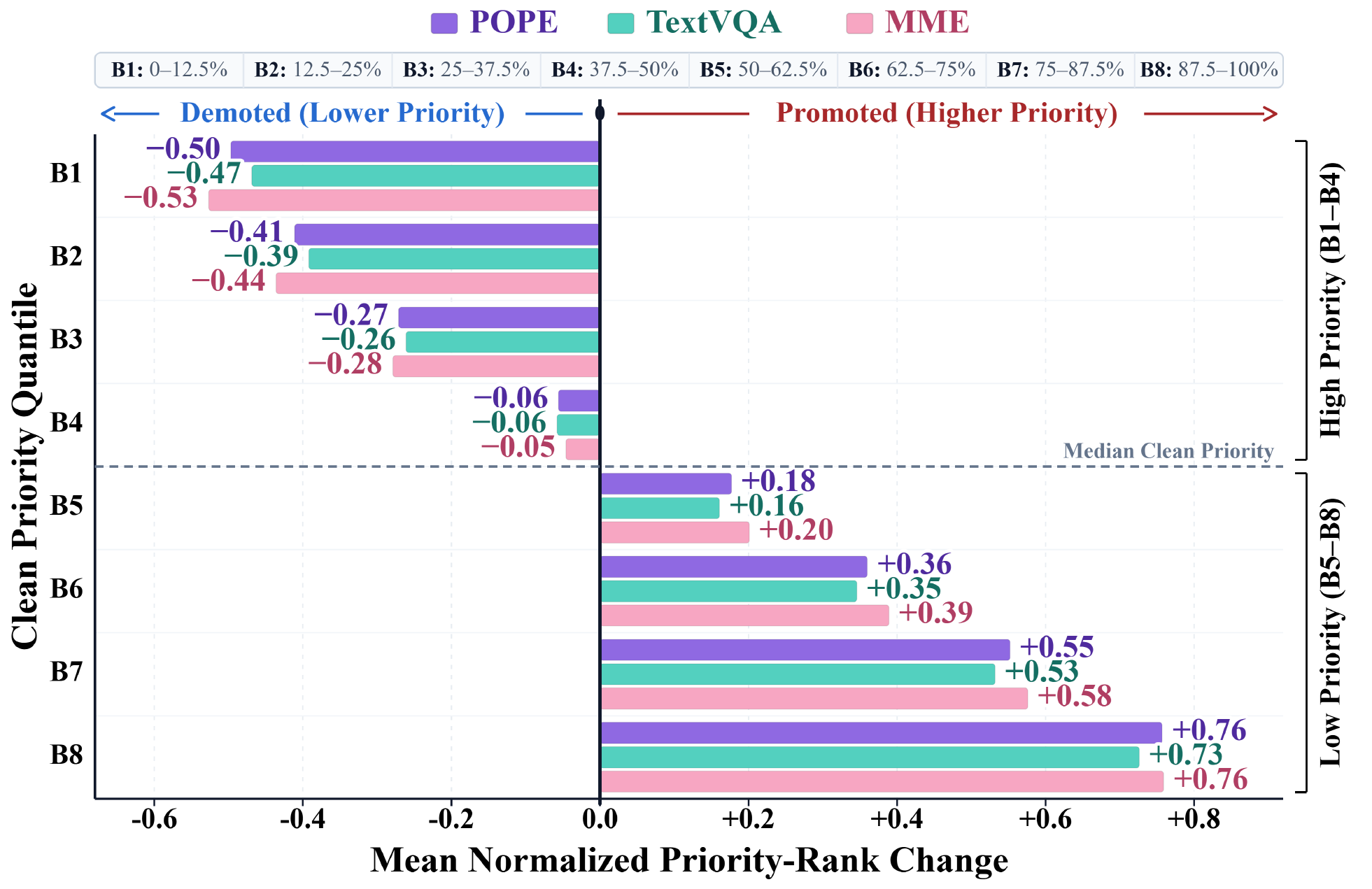}
        \caption{CIRA-induced priority reallocation across the clean visual-token
        ranking at $K=128$.}
        \label{fig:rank_reallocation}
    \end{minipage}
    \hfill
    \begin{minipage}[t]{0.40\textwidth}
        \centering
        \includegraphics[width=\linewidth]{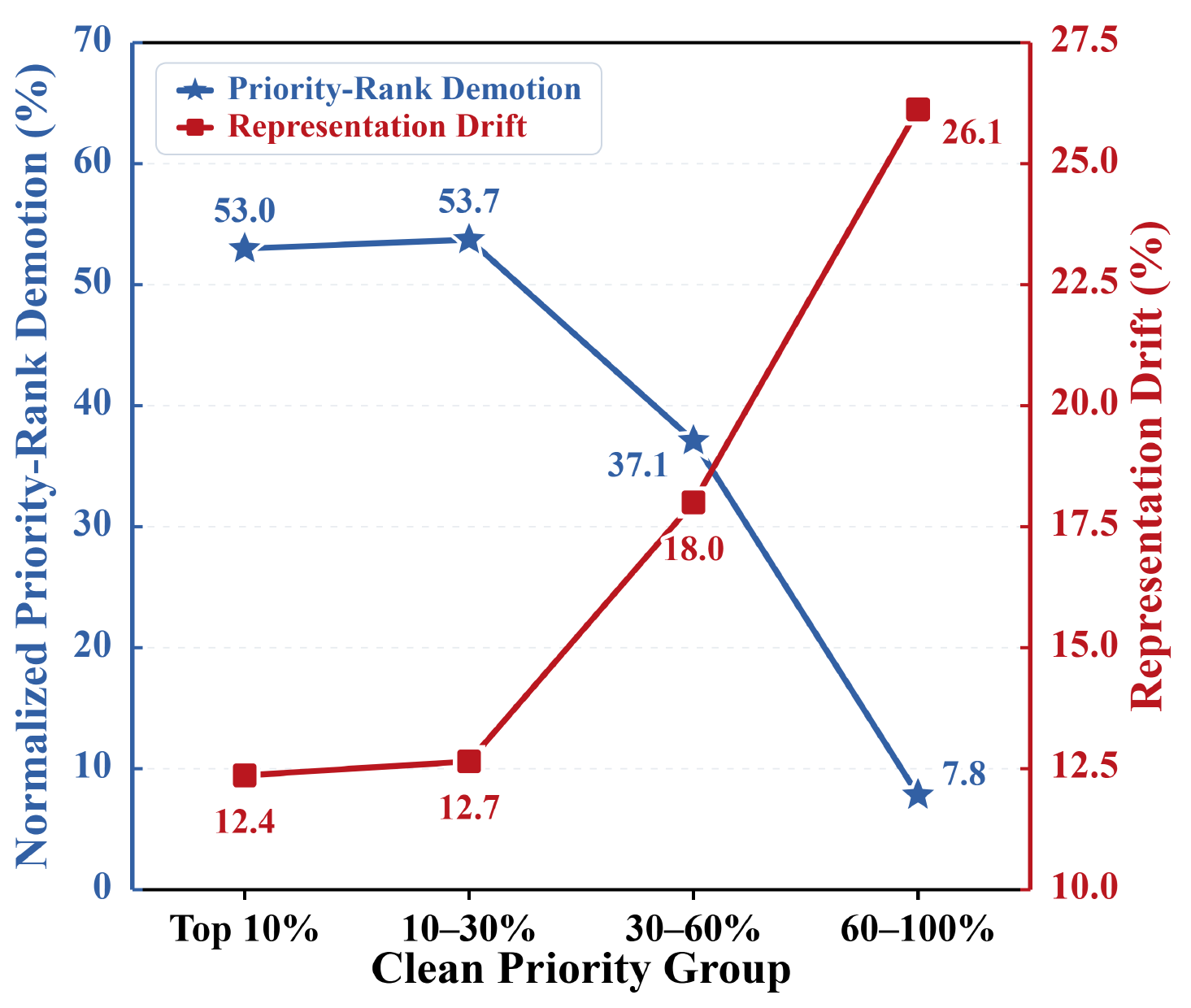}
        \caption{Priority reallocation and representation drift across token groups.}
        \label{fig:selection_representation_decoupling}
    \end{minipage}
\end{figure*}

\paragraph{Priority reallocation.}
Figure~\ref{fig:rank_reallocation} shows a near-monotonic priority
reallocation across all three benchmarks: the highest-priority octile is
demoted by $0.47$--$0.53$ normalized-rank units, whereas the lowest-priority
octile is promoted by $0.73$--$0.76$, with the sign changing near the median.
This cross-quantile pattern is not confined to a single retention boundary.

\paragraph{Priority reallocation versus representation drift.}
Figure~\ref{fig:selection_representation_decoupling} shows that rank
displacement and representation drift behave differently across clean-priority
groups. Rank demotion peaks at $53.0\%$ and $53.7\%$ in the top
$10\%$ and $10$--$30\%$ groups, where representation drift is lowest at
$12.4\%$ and $12.7\%$. The lowest-priority $60$--$100\%$ group shows the
reverse pattern, with $7.8\%$ rank demotion and $26.1\%$ drift. This separation
indicates that CIRA reallocates clean high-priority tokens while limiting their
representation drift, consistent with the HEP objective.

\paragraph{Component ablation.}
On VisionZip, Table~\ref{tab:ablation} isolates the two objectives. Removing
GSH reduces CSFR averaged across the three benchmarks and four budgets from
$18.22\%$ to $3.68\%$. Removing HEP leaves this mean nearly unchanged
($17.04\%$) but raises mean Full ASR from $6.92\%$ to $23.10\%$; replacing it
with global representation preservation yields $15.71\%$ mean CSFR and
$8.66\%$ Full ASR. Global preservation yields lower CSFR and higher Full ASR than HEP, supporting preservation focused on displaced evidence rather than a uniform constraint.

\paragraph{\TakeawayMark\ \textbf{Takeaway.}}
GSH drives compression-specific failure induction, while HEP limits
representation drift in displaced evidence and substantially reduces full-token
degradation.

\input{tables/objective_ablation_tcs_layout}

%% file: tables/main_results.tex
\begin{table}[t]
\centering
\scriptsize

\caption{
CSFR and Full ASR on LLaVA-v1.5-7B. For downstream-agnostic attacks,
\protect\colorbox{dangercolor}{\textcolor{dangertextcolor}{\textbf{red}}} and
\protect\colorbox{safecolor}{\textcolor{safetextcolor}{\textbf{blue}}} cells
mark highest CSFR and lowest Full ASR, and underlining marks second best;
CAA$^{\dagger}$ uses the downstream question and language model
during optimization and is excluded from the markings.
}
\label{tab:results_csfr}
\vspace{2pt}

\begin{adjustbox}{width=\textwidth}
\renewcommand{\arraystretch}{1.13}
\setlength{\tabcolsep}{3pt}

\begin{tabular}{
>{\centering\arraybackslash}m{2.05cm}
!{\color{black}\vrule width 0.45pt}
cccc
!{\color{black}\vrule width 0.45pt}
cccc
!{\color{black}\vrule width 0.45pt}
cccc
}

\Xhline{2pt}

\rowcolor{headercolor}
\textbf{Metric}
& \multicolumn{4}{c!{\color{black}\vrule width 0.45pt}}{\textbf{POPE}}
& \multicolumn{4}{c!{\color{black}\vrule width 0.45pt}}{\textbf{TextVQA}}
& \multicolumn{4}{c}{\textbf{MME}}
\\

\rowcolor{headercolor}
{}
& \textbf{VEAttack} & \textbf{CAGE} & \textbf{CAA$^\dagger$} & \textbf{CIRA}
& \textbf{VEAttack} & \textbf{CAGE} & \textbf{CAA$^\dagger$} & \textbf{CIRA}
& \textbf{VEAttack} & \textbf{CAGE} & \textbf{CAA$^\dagger$} & \textbf{CIRA}
\\

\Xhline{2pt}

% Full-token inference
\multicolumn{13}{>{\columncolor{gray!20}}c}{
\textbf{Full-token Inference (Compressor-independent)}
}
\\
\hline

\textbf{Full ASR}\,\textcolor{blue}{$\downarrow$}
& \secondbest{38.65} & 39.48 & 2.72 & \safecell{4.96}
& \secondbest{60.95} & 71.35 & 3.82 & \safecell{10.61}
& \secondbest{38.26} & 39.77 & 3.41 & \safecell{5.18}
\\

\Xhline{1.5pt}

% VisionZip
\multicolumn{13}{>{\columncolor{gray!20}}c}{
\textbf{VisionZip}
}
\\
\hline

\textbf{CSFR@192}\,\textcolor{red}{$\uparrow$}
& \secondbest{4.05} & 3.19 & 1.72 & \dangercell{7.61}
& \secondbest{4.87} & 2.92 & 2.14 & \dangercell{11.35}
& 2.83 & \secondbest{4.04} & 1.62 & \dangercell{7.40}
\\

\textbf{CSFR@128}\,\textcolor{red}{$\uparrow$}
& 4.98 & \secondbest{5.72} & 2.87 & \dangercell{14.05}
& \secondbest{5.32} & \secondbest{5.32} & 2.05 & \dangercell{18.21}
& 3.33 & \secondbest{5.55} & 1.80 & \dangercell{10.26}
\\

\textbf{CSFR@64}\,\textcolor{red}{$\uparrow$}
& 5.05 & \secondbest{9.71} & 3.46 & \dangercell{22.47}
& \secondbest{6.28} & \secondbest{6.28} & 2.39 & \dangercell{27.41}
& 3.64 & \secondbest{5.97} & 2.92 & \dangercell{19.07}
\\

\textbf{CSFR@32}\,\textcolor{red}{$\uparrow$}
& 6.52 & \secondbest{9.48} & 4.57 & \dangercell{26.96}
& \secondbest{9.76} & 5.24 & 5.01 & \dangercell{31.70}
& 4.56 & \secondbest{5.19} & 2.86 & \dangercell{22.17}
\\

\textbf{Avg. CSFR}\,\textcolor{red}{$\uparrow$}
& 5.15 & \secondbest{7.03} & 3.16 & \dangercell{17.77}
& \secondbest{6.56} & 4.94 & 2.90 & \dangercell{22.17}
& 3.59 & \secondbest{5.19} & 2.30 & \dangercell{14.73}
\\

\Xhline{1.5pt}

% VisPruner
\multicolumn{13}{>{\columncolor{gray!20}}c}{
\textbf{VisPruner}
}
\\
\hline

\textbf{CSFR@192}\,\textcolor{red}{$\uparrow$}
& 3.42 & \secondbest{3.79} & 1.23 & \dangercell{5.87}
& 2.92 & \secondbest{3.50} & 1.75 & \dangercell{10.50}
& 2.56 & \secondbest{3.50} & 1.22 & \dangercell{7.14}
\\

\textbf{CSFR@128}\,\textcolor{red}{$\uparrow$}
& \secondbest{4.60} & 4.35 & 2.97 & \dangercell{9.07}
& \secondbest{3.80} & 3.40 & 2.18 & \dangercell{13.18}
& 2.88 & \secondbest{4.79} & 2.33 & \dangercell{11.23}
\\

\textbf{CSFR@64}\,\textcolor{red}{$\uparrow$}
& 6.27 & \secondbest{7.45} & 4.68 & \dangercell{15.29}
& \secondbest{5.89} & 5.47 & 4.08 & \dangercell{19.13}
& 3.33 & \secondbest{5.36} & 4.47 & \dangercell{18.70}
\\

\textbf{CSFR@32}\,\textcolor{red}{$\uparrow$}
& 7.59 & \secondbest{10.46} & 8.29 & \dangercell{22.21}
& \secondbest{8.07} & 7.33 & 4.63 & \dangercell{25.57}
& 4.13 & \secondbest{5.66} & 6.73 & \dangercell{21.87}
\\

\textbf{Avg. CSFR}\,\textcolor{red}{$\uparrow$}
& 5.47 & \secondbest{6.51} & 4.29 & \dangercell{13.11}
& \secondbest{5.17} & 4.93 & 3.16 & \dangercell{17.09}
& 3.22 & \secondbest{4.83} & 3.69 & \dangercell{14.73}
\\

\Xhline{1.5pt}

% PruMerge
\multicolumn{13}{>{\columncolor{gray!20}}c}{
\textbf{PruMerge}
}
\\
\hline

\textbf{CSFR@192}\,\textcolor{red}{$\uparrow$}
& \secondbest{5.33} & 3.65 & 2.66 & \dangercell{22.86}
& \secondbest{6.02} & 4.86 & 3.44 & \dangercell{28.66}
& 2.64 & \secondbest{4.69} & 1.75 & \dangercell{20.64}
\\

\textbf{CSFR@128}\,\textcolor{red}{$\uparrow$}
& 5.39 & \secondbest{5.54} & 2.92 & \dangercell{29.59}
& \secondbest{6.51} & 5.35 & 4.86 & \dangercell{36.23}
& 3.62 & \secondbest{5.28} & 3.02 & \dangercell{22.93}
\\

\textbf{CSFR@64}\,\textcolor{red}{$\uparrow$}
& 6.96 & \secondbest{9.38} & 3.63 & \dangercell{32.53}
& \secondbest{8.29} & 6.64 & 4.49 & \dangercell{43.91}
& 3.50 & \secondbest{5.78} & 3.04 & \dangercell{29.98}
\\

\textbf{CSFR@32}\,\textcolor{red}{$\uparrow$}
& 9.40 & \secondbest{11.76} & 3.44 & \dangercell{32.29}
& \secondbest{9.18} & 6.70 & 5.20 & \dangercell{48.68}
& 4.06 & \secondbest{5.78} & 2.81 & \dangercell{35.16}
\\

\textbf{Avg. CSFR}\,\textcolor{red}{$\uparrow$}
& 6.77 & \secondbest{7.58} & 3.16 & \dangercell{29.32}
& \secondbest{7.50} & 5.89 & 4.50 & \dangercell{39.37}
& 3.45 & \secondbest{5.38} & 2.66 & \dangercell{27.18}
\\

\Xhline{1.5pt}

% FastV
\multicolumn{13}{>{\columncolor{gray!20}}c}{
\textbf{FastV}
}
\\
\hline

\textbf{CSFR@192}\,\textcolor{red}{$\uparrow$}
& 4.61 & \secondbest{4.74} & 2.30 & \dangercell{7.04}
& \secondbest{4.29} & 4.09 & 1.43 & \dangercell{9.61}
& 2.77 & \secondbest{2.90} & 1.38 & \dangercell{5.81}
\\

\textbf{CSFR@128}\,\textcolor{red}{$\uparrow$}
& \secondbest{6.85} & 6.59 & 2.90 & \dangercell{11.46}
& \secondbest{7.10} & 5.38 & 2.15 & \dangercell{13.55}
& 3.17 & \secondbest{3.89} & 1.44 & \dangercell{7.64}
\\

\textbf{CSFR@64}\,\textcolor{red}{$\uparrow$}
& 9.90 & \secondbest{10.47} & 3.44 & \dangercell{22.24}
& \secondbest{9.11} & 7.67 & 4.08 & \dangercell{23.74}
& 3.65 & \secondbest{5.18} & 2.89 & \dangercell{15.07}
\\

\textbf{CSFR@32}\,\textcolor{red}{$\uparrow$}
& \secondbest{12.32} & 10.40 & 4.48 & \dangercell{25.44}
& 7.80 & \secondbest{8.38} & 3.76 & \dangercell{32.95}
& 4.24 & \secondbest{5.55} & 6.20 & \dangercell{20.39}
\\

\textbf{Avg. CSFR}\,\textcolor{red}{$\uparrow$}
& \secondbest{8.42} & 8.05 & 3.28 & \dangercell{16.55}
& \secondbest{7.08} & 6.38 & 2.85 & \dangercell{19.96}
& 3.46 & \secondbest{4.38} & 2.98 & \dangercell{12.23}
\\

\Xhline{2pt}

\end{tabular}
\end{adjustbox}
\end{table}

%% file: tables/objective_ablation_tcs_layout.tex
\begin{table*}[t!]
\centering

\input{tables/objective_ablation}\hfill%
\input{tables/tcs_defense}
\vspace{-3pt}
\end{table*}

%% file: tables/objective_ablation.tex
\begin{minipage}[t]{0.452\textwidth}
\vspace{0pt}
\centering

\caption{CIRA ablation on \mbox{VisionZip}; arrows show changes from full CIRA.}
\label{tab:ablation}

\vspace{2pt}

\scriptsize
\renewcommand{\arraystretch}{1.04}
\setlength{\tabcolsep}{2.2pt}

\begin{adjustbox}{width=\linewidth}
\begin{tabular}{
>{\centering\arraybackslash}m{1.48cm}
!{\color{black}\vrule width 0.45pt}
c
!{\color{black}\vrule width 0.45pt}
c
!{\color{black}\vrule width 0.45pt}
cc
}

\Xhline{2pt}

\rowcolor{headercolor}
{}
& \multicolumn{1}{
    c!{\color{black}\vrule width 0.45pt}
  }{\textbf{Full}}
& \multicolumn{1}{
    c!{\color{black}\vrule width 0.45pt}
  }{\textbf{Selection}}
& \multicolumn{2}{c}{\textbf{Preservation}}
\\

\rowcolor{headercolor}
\textbf{Metric}
& \textbf{CIRA}
& \textbf{w/o Sel.}
& \textbf{w/o Pres.}
& \textbf{Global Pres.}
\\

\Xhline{2pt}

% POPE
\multicolumn{5}{>{\columncolor{black!12}}c}{
\textbf{POPE}
}
\\
\Xhline{0.45pt}

\textbf{Full ASR}\,\textcolor{blue}{$\downarrow$}
& \textbf{4.96}
& 2.01\AblDown{2.95}
& 15.72\AblUp{10.76}
& 4.73\AblDown{0.23}
\\

\Xhline{0.45pt}

\textbf{CSFR@192}\,\textcolor{red}{$\uparrow$}
& \textbf{7.61}
& 1.60\AblDown{6.01}
& 9.94\AblUp{2.33}
& 4.79\AblDown{2.82}
\\

\textbf{CSFR@128}\,\textcolor{red}{$\uparrow$}
& \textbf{14.05}
& 1.99\AblDown{12.06}
& 15.42\AblUp{1.37}
& 9.95\AblDown{4.10}
\\

\textbf{CSFR@64}\,\textcolor{red}{$\uparrow$}
& \textbf{22.47}
& 4.39\AblDown{18.08}
& 22.07\AblDown{0.40}
& 19.95\AblDown{2.52}
\\

\textbf{CSFR@32}\,\textcolor{red}{$\uparrow$}
& \textbf{26.96}
& 5.63\AblDown{21.33}
& 26.81\AblDown{0.15}
& 26.37\AblDown{0.59}
\\

\textbf{Avg. CSFR}\,\textcolor{red}{$\uparrow$}
& \textbf{17.77}
& 3.40\AblDown{14.37}
& 18.56\AblUp{0.79}
& 15.26\AblDown{2.51}
\\

\Xhline{1.2pt}

% TextVQA
\multicolumn{5}{>{\columncolor{black!12}}c}{
\textbf{TextVQA}
}
\\
\Xhline{0.45pt}

\textbf{Full ASR}\,\textcolor{blue}{$\downarrow$}
& \textbf{10.61}
& 6.91\AblDown{3.70}
& 36.00\AblUp{25.39}
& 15.82\AblUp{5.20}
\\

\Xhline{0.45pt}

\textbf{CSFR@192}\,\textcolor{red}{$\uparrow$}
& \textbf{11.35}
& 3.90\AblDown{7.45}
& 9.94\AblDown{1.41}
& 8.58\AblDown{2.77}
\\

\textbf{CSFR@128}\,\textcolor{red}{$\uparrow$}
& \textbf{18.21}
& 2.05\AblDown{16.16}
& 15.20\AblDown{3.02}
& 12.53\AblDown{5.69}
\\

\textbf{CSFR@64}\,\textcolor{red}{$\uparrow$}
& \textbf{27.41}
& 5.43\AblDown{21.98}
& 23.70\AblDown{3.72}
& 23.04\AblDown{4.37}
\\

\textbf{CSFR@32}\,\textcolor{red}{$\uparrow$}
& \textbf{31.70}
& 6.44\AblDown{25.26}
& 26.73\AblDown{4.97}
& 36.28\AblUp{4.57}
\\

\textbf{Avg. CSFR}\,\textcolor{red}{$\uparrow$}
& \textbf{22.17}
& 4.46\AblDown{17.71}
& 18.89\AblDown{3.28}
& 20.11\AblDown{2.06}
\\

\Xhline{1.2pt}

% MME
\multicolumn{5}{>{\columncolor{black!12}}c}{
\textbf{MME}
}
\\
\Xhline{0.45pt}

\textbf{Full ASR}\,\textcolor{blue}{$\downarrow$}
& \textbf{5.18}
& 2.40\AblDown{2.77}
& 17.57\AblUp{12.40}
& 5.44\AblUp{0.26}
\\

\Xhline{0.45pt}

\textbf{CSFR@192}\,\textcolor{red}{$\uparrow$}
& \textbf{7.40}
& 1.62\AblDown{5.79}
& 7.55\AblUp{0.14}
& 5.80\AblDown{1.61}
\\

\textbf{CSFR@128}\,\textcolor{red}{$\uparrow$}
& \textbf{10.26}
& 1.93\AblDown{8.33}
& 11.33\AblUp{1.06}
& 7.73\AblDown{2.53}
\\

\textbf{CSFR@64}\,\textcolor{red}{$\uparrow$}
& \textbf{19.07}
& 4.09\AblDown{14.98}
& 17.23\AblDown{1.84}
& 14.45\AblDown{4.62}
\\

\textbf{CSFR@32}\,\textcolor{red}{$\uparrow$}
& \textbf{22.17}
& 5.08\AblDown{17.09}
& 18.57\AblDown{3.60}
& 19.05\AblDown{3.12}
\\

\textbf{Avg. CSFR}\,\textcolor{red}{$\uparrow$}
& \textbf{14.73}
& 3.18\AblDown{11.55}
& 13.67\AblDown{1.06}
& 11.76\AblDown{2.97}
\\

\Xhline{2pt}

\end{tabular}
\end{adjustbox}

\end{minipage}%

%% file: tables/tcs_defense.tex
\begin{minipage}[t]{0.538\textwidth}
\vspace{0pt}
\centering

\caption{TCS evaluation on VisionZip; arrows denote changes from matched no-defense
baselines.}
\label{tab:tcs_defense}

\vspace{2pt}

\scriptsize
\renewcommand{\arraystretch}{1.04}
\setlength{\tabcolsep}{2.5pt}

\begin{adjustbox}{width=\linewidth}
\begin{tabular}{
>{\centering\arraybackslash}m{1.52cm}
!{\color{black}\vrule width 0.45pt}
cc
!{\color{black}\vrule width 0.45pt}
cc
!{\color{black}\vrule width 0.45pt}
cc
}

\Xhline{2pt}

\rowcolor{headercolor}
{}
& \multicolumn{2}{
    c!{\color{black}\vrule width 0.45pt}
  }{\textbf{CIRA}}
& \multicolumn{2}{
    c!{\color{black}\vrule width 0.45pt}
  }{\textbf{CAGE}}
& \multicolumn{2}{c}{\textbf{Adaptive CIRA}}
\\

\rowcolor{headercolor}
\textbf{Metric}
& \textbf{None}
& \textbf{TCS}
& \textbf{None}
& \textbf{TCS}
& \textbf{None}
& \textbf{TCS}
\\

\Xhline{2pt}

% POPE
\multicolumn{7}{>{\columncolor{black!12}}c}{
\textbf{POPE}
}
\\
\Xhline{0.45pt}

\textbf{Full ASR}
& \multicolumn{2}{
    c!{\color{black}\vrule width 0.45pt}
  }{4.96}
& \multicolumn{2}{
    c!{\color{black}\vrule width 0.45pt}
  }{39.48}
& \multicolumn{2}{c}{3.78}
\\

\Xhline{0.45pt}

\textbf{CSFR@192}\,\textcolor{blue}{$\downarrow$}
& 7.61
& 2.60\DefenseDown{5.01}
& 3.19
& 4.34\DefenseUp{1.15}
& 6.99
& 6.57\DefenseDown{0.42}
\\

\textbf{CSFR@128}\,\textcolor{blue}{$\downarrow$}
& 14.05
& 3.40\DefenseDown{10.65}
& 5.72
& 4.79\DefenseDown{0.93}
& 10.07
& 10.71\DefenseUp{0.64}
\\

\textbf{CSFR@64}\,\textcolor{blue}{$\downarrow$}
& 22.47
& 2.84\DefenseDown{19.63}
& 9.71
& 5.68\DefenseDown{4.03}
& 20.08
& 13.65\DefenseDown{6.43}
\\

\textbf{CSFR@32}\,\textcolor{blue}{$\downarrow$}
& 26.96
& 4.21\DefenseDown{22.75}
& 9.48
& 5.52\DefenseDown{3.96}
& 28.00
& 21.92\DefenseDown{6.08}
\\

\textbf{Avg. CSFR}\,\textcolor{blue}{$\downarrow$}
& 17.77
& 3.26\DefenseDown{14.51}
& 7.03
& 5.08\DefenseDown{1.95}
& 16.29
& 13.21\DefenseDown{3.08}
\\

\Xhline{1.2pt}

% TextVQA
\multicolumn{7}{>{\columncolor{black!12}}c}{
\textbf{TextVQA}
}
\\
\Xhline{0.45pt}

\textbf{Full ASR}
& \multicolumn{2}{
    c!{\color{black}\vrule width 0.45pt}
  }{10.61}
& \multicolumn{2}{
    c!{\color{black}\vrule width 0.45pt}
  }{71.35}
& \multicolumn{2}{c}{11.45}
\\

\Xhline{0.45pt}

\textbf{CSFR@192}\,\textcolor{blue}{$\downarrow$}
& 11.35
& 2.39\DefenseDown{8.96}
& 2.92
& 1.59\DefenseDown{1.33}
& 8.38
& 8.17\DefenseDown{0.21}
\\

\textbf{CSFR@128}\,\textcolor{blue}{$\downarrow$}
& 18.21
& 2.07\DefenseDown{16.14}
& 5.32
& 2.48\DefenseDown{2.84}
& 11.09
& 9.30\DefenseDown{1.79}
\\

\textbf{CSFR@64}\,\textcolor{blue}{$\downarrow$}
& 27.41
& 4.50\DefenseDown{22.91}
& 6.28
& 3.43\DefenseDown{2.85}
& 26.09
& 14.13\DefenseDown{11.95}
\\

\textbf{CSFR@32}\,\textcolor{blue}{$\downarrow$}
& 31.70
& 3.09\DefenseDown{28.61}
& 5.24
& 4.99\DefenseDown{0.25}
& 38.90
& 25.65\DefenseDown{13.25}
\\

\textbf{Avg. CSFR}\,\textcolor{blue}{$\downarrow$}
& 22.17
& 3.01\DefenseDown{19.16}
& 4.94
& 3.12\DefenseDown{1.82}
& 21.11
& 14.31\DefenseDown{6.80}
\\

\Xhline{1.2pt}

% MME
\multicolumn{7}{>{\columncolor{black!12}}c}{
\textbf{MME}
}
\\
\Xhline{0.45pt}

\textbf{Full ASR}
& \multicolumn{2}{
    c!{\color{black}\vrule width 0.45pt}
  }{5.18}
& \multicolumn{2}{
    c!{\color{black}\vrule width 0.45pt}
  }{39.77}
& \multicolumn{2}{c}{6.57}
\\

\Xhline{0.45pt}

\textbf{CSFR@192}\,\textcolor{blue}{$\downarrow$}
& 7.40
& 2.72\DefenseDown{4.68}
& 4.04
& 2.31\DefenseDown{1.73}
& 3.64
& 5.03\DefenseUp{1.39}
\\

\textbf{CSFR@128}\,\textcolor{blue}{$\downarrow$}
& 10.26
& 2.64\DefenseDown{7.62}
& 5.55
& 2.50\DefenseDown{3.05}
& 9.53
& 7.22\DefenseDown{2.31}
\\

\textbf{CSFR@64}\,\textcolor{blue}{$\downarrow$}
& 19.07
& 4.53\DefenseDown{14.54}
& 5.97
& 4.09\DefenseDown{1.88}
& 18.54
& 11.68\DefenseDown{6.86}
\\

\textbf{CSFR@32}\,\textcolor{blue}{$\downarrow$}
& 22.17
& 5.75\DefenseDown{16.42}
& 5.19
& 4.67\DefenseDown{0.52}
& 23.17
& 18.35\DefenseDown{4.82}
\\

\textbf{Avg. CSFR}\,\textcolor{blue}{$\downarrow$}
& 14.73
& 3.91\DefenseDown{10.82}
& 5.19
& 3.39\DefenseDown{1.80}
& 13.72
& 10.57\DefenseDown{3.15}
\\

\Xhline{2pt}

\end{tabular}
\end{adjustbox}

\end{minipage}%

%% file: 7-defense.tex
\section{Selection Stabilization Defense}
\label{sec:defense}

To test whether stabilizing token selection can suppress CIRA, TCS exploits
cross-view priority stability: clean high-priority evidence tends to remain
stable under small translations, whereas attack-induced replacements are more
view-sensitive. At the LLaVA--VisionZip token-priority interface, TCS uses four
translated views
$\mathcal{V}=\{T_{0,0},T_{d,0},T_{0,d},T_{d,d}\}$ with $d=7$ pixels. For view
$v$, $\mathbf{s}^{(v)}$ is its priority-score vector, $\mathcal{A}_{v}$ maps the
translated score grid back to the reference coordinates, and rank $1$ denotes
the highest priority. TCS selects the $K$ tokens with highest aligned
rank-quantile consensus:
\begin{equation}
q_i^{(v)}
=1-\frac{r_i\!\left(\mathcal{A}_{v}(\mathbf{s}^{(v)})\right)-1}{N-1},
\quad
\bar q_i
=\frac{1}{|\mathcal{V}|}\sum_{v\in\mathcal{V}}q_i^{(v)},
\quad
\mathcal{I}^{\mathrm{TCS}}_{K}
=\operatorname{TopK}(\bar{\mathbf q},K).
\label{eq:tcs_main}
\end{equation}
Rank quantiles make priorities comparable across views despite differences in
score scale. The selected indices are applied to the unshifted-view features,
leaving aggregation and full-token inference unchanged. Construction and
mechanism details appear in \AppendixRef{sec:appendix_defense}; complementary
utility results are reported in \AppendixRef{sec:appendix_task_utility}.

\paragraph{Defense effectiveness.}
Within the matched evaluation blocks in Table~\ref{tab:tcs_defense}, TCS
reduces standard CIRA CSFR averaged across the three datasets and four budgets
from $18.22\%$ to $3.39\%$, an $81.4\%$ relative reduction, compared with a
$32.5\%$ reduction for CAGE. Suppression strengthens as the compression budget
decreases, with the largest reductions under tighter compression.

\paragraph{Adaptive stress test.}
Because TCS is deterministic and public, we also evaluate an adaptive attacker
that optimizes the CIRA objectives over the same four views while sharing one
image-space perturbation:
\begin{equation}
\mathcal{L}_{\mathrm{adapt}}(\delta)
=\frac{1}{|\mathcal{V}|}
\sum_{v\in\mathcal{V}}
\left[
\mathcal{L}_{\mathrm{GSH}}^{(v)}(\delta)
+\lambda\mathcal{L}_{\mathrm{HEP}}^{(v)}(\delta)
\right].
\label{eq:adaptive_multiview}
\end{equation}
Adaptive CIRA uses the same access assumptions and perturbation budget as
standard CIRA. Under TCS, its mean CSFR rises from $3.39\%$ to $12.70\%$,
reaching $74.5\%$ of Adaptive CIRA's matched undefended value ($17.04\%$).
TCS therefore retains a smaller but nonzero effect against the adaptive attack.
Additional optimization and mechanism diagnostics appear in
\AppendixRef{sec:appendix_adaptive}.

%% file: 8-conclution.tex
\section{Conclusion}
\label{sec:conclusion}

Visual-token compression changes not only inference cost but also which visual
evidence remains available after compression. By pairing full-token and
compressed inference on the same adversarial input, we attribute failures
specifically to the compression path. CIRA induces such failures across unknown
compressor and budget settings by reallocating token priorities while limiting
full-token degradation. Controlled diagnostics further show that retained-set
allocation affects compressed correctness and that recovery is negatively
associated with representation drift in displaced evidence. Selection
stabilization substantially suppresses CIRA, although adaptive optimization
partially restores its effectiveness. Together, these results support treating
the compression boundary as a security-relevant component of LVLM deployment
and motivate paired robustness evaluation for compressed LVLMs.

\section*{AI Use Statement}
Generative AI tools were used in a limited supporting role for this work,
including language editing and polishing, literature retrieval and discovery,
research ideation and technical execution support, and drafting and revising
parts of the manuscript. In particular, these tools were used to improve the
clarity and fluency of writing, help identify related literature, provide
technical suggestions for coding and experimental workflows, and assist in
refining sections of the paper during revision. All AI-assisted content was
independently checked, validated, and, where necessary, modified by the
authors. The authors retain full responsibility for the scientific content,
methodological choices, experimental evidence, interpretations, and final form
of the manuscript.

\section*{Reproducibility Statement}
We provide comprehensive details to facilitate the reproduction and verification
of CIRA. The threat model, compression-specific failure formulation, and attack
objectives are described in Sections~\ref{sec:motivation} and
~\ref{sec:methodology}, including the Global Selection Hijacking (GSH) and
Hidden-Evidence Preservation (HEP) objectives and their joint optimization. The
optimization procedure and experimental details are provided in
Appendices~\ref{sec:appendix_cira} and
\ref{sec:appendix_detailed_setup}, respectively. Main results, mechanistic
analyses, and defense evaluations are reported in
Sections~\ref{sec:main_results}, \ref{sec:mechanism_analysis_exp}, and
\ref{sec:defense}. Additional analyses, including retained-set diagnostics,
sensitivity studies, cross-model evaluation, task-utility results, limitations,
and qualitative cases, are provided in Appendices~\ref{sec:appendix_motivation},
\ref{sec:appendix_sensitivity},
\ref{sec:appendix_cross_model},
\ref{sec:appendix_task_utility},
\ref{sec:appendix_limitations}, and
\ref{sec:case-studies}. Code and scripts for reproducing the reported
experiments are provided in the Github repository.

\section*{Ethics Statement}
This work investigates the adversarial robustness of LVLMs under visual-token
compression in a controlled research setting. Our experiments use publicly
available benchmark datasets (e.g., POPE, TextVQA, and MME) and publicly released
pretrained models (e.g., LLaVA-v1.5-7B, Qwen3-VL-8B-Instruct, and
InternVL3.5-8B), involving no human subjects or personally identifiable
information. We recognize the dual-use nature of adversarial robustness
research. CIRA is developed to identify compression-specific vulnerabilities
and support robustness evaluation and defense development. The proposed attack
is evaluated under a restricted threat model with bounded image-space
perturbations and vision-encoder-only access. We transparently report the attack
assumptions and evaluation protocols to facilitate reproducibility and
responsible security research.

%% file: 9-appendix.tex
\setcounter{dbltopnumber}{4}
\renewcommand{\dbltopfraction}{0.90}
\setlength{\dblfloatsep}{8pt plus 2pt minus 2pt}

\section*{Appendix Contents}

\begingroup
\hypersetup{linkcolor=black}
\setlength{\parindent}{0pt}
\setlength{\parskip}{2pt}
\newcommand{\AppendixContentsLine}[2]{%
  \noindent
  \hyperref[#1]{%
    \makebox[\linewidth][l]{%
      \textbf{\underline{Appendix~\textcolor{red}{\ref*{#1}}}}.~%
      #2\nobreak\dotfill\nobreak\pageref*{#1}%
    }%
  }\par
}
\AppendixContentsLine{sec:appendix_cira}{CIRA Optimization Procedure}
\AppendixContentsLine{sec:appendix_detailed_setup}{Detailed Experimental Setup}
\AppendixContentsLine{sec:appendix_motivation}{Controlled Retained-Set Allocation Diagnostics}
\AppendixContentsLine{sec:appendix_sensitivity}{Sensitivity Analyses}
\AppendixContentsLine{sec:appendix_cross_model}{Cross-Model Scope}
\AppendixContentsLine{sec:appendix_defense}{Selection Stabilization Defense}
\AppendixContentsLine{sec:appendix_task_utility}{Complementary Task-Utility Results}
\AppendixContentsLine{sec:appendix_limitations}{Limitation Discussion and Future Work}
\AppendixContentsLine{sec:case-studies}{Qualitative Case Studies}
\endgroup

\input{appendix/implementation}

\input{appendix/detailed_setup}

\input{appendix/diagnostics}

\input{appendix/sensitivity}

\input{appendix/cross_model}

\input{appendix/defense}

\input{appendix/utility}

\input{appendix/limitations}

\input{appendix/case_studies}

%% file: appendix/implementation.tex
\section{CIRA Optimization Procedure}
\label{sec:appendix_cira}

Algorithm~\ref{alg:cira} gives the image-space optimization induced by
CIRA's two encoder-side objectives. Clean features and priority scores are
cached once; adversarial scores, ranks, and displacement weights are refreshed
at every step. The priority rule $\mathcal P$ maps encoder attention to the
token-score vector used by GSH.

\begin{algorithm}[H]
\caption{Projected sign-gradient optimization of CIRA.}
\label{alg:cira}
\footnotesize
\begin{algorithmic}[1]
\Require Image $x$; encoder $\mathcal{E}$; priority rule $\mathcal{P}$
\Require Candidate interval $[K_{\min},K_{\max}]$; $\epsilon,\epsilon_s,\epsilon_h$
\Require Step size $\alpha$; steps $T$; preservation weight $\lambda$
\Ensure Adversarial image $x^{\mathrm{adv}}$
\State $\mathcal{K}\gets\{K_{\min},\ldots,K_{\max}\}$
\State $(\mathbf{H}^{c},\mathbf{s}^{c})\gets
(\mathcal{E}(x),\mathcal{P}(x))$
\State $\mathbf{r}^{c}\gets
\operatorname{Rank}_{\downarrow}(\mathbf{s}^{c})$
\State $\mathbf{u}^{c}\gets\mathcal{R}(\mathbf{s}^{c})$
\State Initialize $\delta^{(0)}\gets\mathbf{0}$
\For{$\tau=0,\ldots,T-1$}
    \State $x^{(\tau)}\gets x+\delta^{(\tau)}$
    \State $(\mathbf{H}^{a},\mathbf{s}^{a})\gets
    (\mathcal{E}(x^{(\tau)}),\mathcal{P}(x^{(\tau)}))$
    \State $\mathcal{L}_{\mathrm{GSH}}\gets
    \operatorname{Align}_{\epsilon_s}(\mathbf{s}^{a},\mathbf{u}^{c})$
    \State $\mathbf{r}^{a}\gets
    \operatorname{Rank}_{\downarrow}(\mathbf{s}^{a})$
    \State $w_i\gets\operatorname{sg}\!\left(
    \Pr_{K\sim\mathrm{Unif}(\mathcal K)}
    [r_i^{c}\leq K<r_i^{a}]\right)$
    \State $d_i\gets\frac{1}{2}
    (1-\cos(\mathbf h_i^c,\mathbf h_i^a))$
    \State $\mathcal{L}_{\mathrm{HEP}}\gets
    -\frac{\sum_i w_i d_i}{\max(\sum_i w_i,\epsilon_h)}$
    \State $\mathcal{L}_{\mathrm{CIRA}}\gets
    \mathcal{L}_{\mathrm{GSH}}+\lambda\mathcal{L}_{\mathrm{HEP}}$
    \State $g\gets\operatorname{sign}
    (\nabla_{\delta}\mathcal{L}_{\mathrm{CIRA}})$
    \State $\delta^{(\tau+1)}\gets
    \Pi_{\Delta_{\epsilon}(x)}(\delta^{(\tau)}+\alpha g)$
\EndFor
\State \Return $x^{\mathrm{adv}}\gets x+\delta^{(T)}$
\end{algorithmic}
\end{algorithm}

Sorting remains outside of the gradient path, $\operatorname{sg}(\cdot)$ blocks
gradients through the discrete weights, and
$\Delta_{\epsilon}(x)=\{\delta:\|\delta\|_{\infty}\leq\epsilon,
x+\delta\in[0,1]^d\}$ is the feasible perturbation set. The update therefore uses only
the vision encoder, priority-scoring rule, and candidate compression-budget
interval.

%% file: appendix/detailed_setup.tex
\section{Detailed Experimental Setup}
\label{sec:appendix_detailed_setup}

\subsection{Models}

\paragraph{LLaVA-v1.5-7B.}
Our primary model is LLaVA-v1.5-7B \citep{liu2024improved} with a CLIP
ViT-L/14-336 vision encoder \citep{radford2021learning}. The released model
feeds 576 penultimate-layer patch tokens to its multimodal projector.
We use LLaVA for the full benchmark--compressor matrix and for the mechanism,
ablation, sensitivity, and defense studies.

\paragraph{Additional LVLM families.}
We additionally evaluate Qwen3-VL-8B-Instruct \citep{bai2025qwen3} and
InternVL3.5-8B \citep{wang2025internvl3}. Qwen3-VL uses its native dynamic
resolution and spatial merger, so its visual-token count varies by image.
InternVL uses one $448\times448$ tile, producing 256 tokens after spatial
downsampling. Accordingly, we use family-specific evaluation budgets of
$\{96,64,32\}$ for Qwen3-VL and $\{128,64,32\}$ for InternVL. CIRA is optimized separately for each model using its native vision encoder.

\subsection{Benchmarks}

We use fixed, randomly sampled 1,000-pair subsets from POPE
\citep{li2023evaluating}, TextVQA \citep{singh2019towards}, and MME
\citep{fu2026mme}. POPE evaluates object hallucination through balanced
object-presence questions; TextVQA tests reading of text in natural images and
provides ten reference answers per question; MME covers perception and cognition
with binary questions. Within each model--benchmark setting, all attacks are evaluated on the same
image--question pairs using a common prompt, decoding protocol, and answer
evaluator. POPE and MME outputs
are lowercased, stripped of punctuation, and matched by the first normalized
yes/no token. TextVQA uses its EvalAI-style normalizer and accepts a match to any
reference answer.

These task-specific rules provide the per-example binary correctness required
by the paired CSF definition. For TextVQA, we use a match to any normalized
reference rather than the benchmark-level soft agreement score; MME is evaluated
per question rather than by its aggregate category score. The same evaluators
are used for clean, full-token, and compressed inference.

\subsection{Compression Mechanisms}

\paragraph{VisionZip.}
VisionZip \citep{yang2025visionzip} retains high-attention tokens and merges the
remainder around uniformly sampled contextual tokens. Compression budgets
$K=32,64,128,192$ use (dominant, contextual) counts $(27,5)$, $(54,10)$,
$(108,20)$, and $(162,30)$, respectively.

\paragraph{VisPruner.}
VisPruner \citep{zhang2025beyond} combines attention-based importance with
feature diversity. Half of each budget is assigned to important tokens and the
remainder to diverse tokens.

\paragraph{PruMerge.}
PruMerge \citep{shang2025llava} selects attention-ranked representatives and
merges nearby tokens by feature similarity. We use $K-1$ representatives
together with one attention-weighted residual aggregate, yielding exactly $K$
output tokens.

\paragraph{FastV.}
FastV \citep{chen2024image} ranks visual tokens using language-model attention
and removes low-ranked tokens after layer 2.

These configurations are used for the primary LLaVA evaluation. For Qwen3-VL
and InternVL, the corresponding compression rules are instantiated on their
native visual-token interfaces while preserving each method's selection and
aggregation principle. Throughout the evaluation, compression budget $K$
denotes the number of post-compression visual tokens passed to subsequent
computation, whether obtained through selection, merging, or both.

\subsection{Implementation}

\paragraph{Inference.}
All experiments run on one NVIDIA GeForce RTX 4090 GPU. Decoding is deterministic
(\texttt{do\_sample=False}) with at most 64 new tokens, using each model's
native image processor and conversation template.

\paragraph{Priority-score instantiation.}
CIRA derives token priorities from model-native late-layer visual attention.
Let $\mathcal{L}_{\mathrm{score}}$ denote the encoder layers used to compute
token priority scores. We use the final, third-to-last, and fifth-to-last
encoder blocks, corresponding to relative layer indices $(-1,-3,-5)$. For the CLIP encoder in the primary setting, the score from
Section~\ref{sec:selection_hijacking} is
\begin{equation}
s_i(x)
=
\frac{1}{|\mathcal{L}_{\mathrm{score}}|}
\sum_{\ell\in\mathcal{L}_{\mathrm{score}}}
\sum_{h=1}^{H}
\alpha_{0,i}^{\ell,h}(x),
\qquad i=1,\ldots,N,
\label{eq:token_importance}
\end{equation}
where $\alpha_{0,i}^{\ell,h}$ is attention from the class token to patch $i$
at head $h$ and visual layer $\ell$, and $H$ is the number of attention heads.

InternVL derives token priorities from class-to-patch visual self-attention and
aggregates them over its spatial-downsampling groups. For Qwen3-VL, which lacks
a class-token routing interface, we use the mean incoming attention over visual
queries and aggregate scores within its native spatial-merger groups. For both
models, priorities are averaged over the same three relative encoder layers to
produce one score per downstream visual token.

\paragraph{Preservation features.}
For LLaVA, HEP measures tokenwise cosine distance between final CLIP patch
features after post-layer normalization. For Qwen3-VL, it averages the
tokenwise distances of the merged main visual features and the DeepStack
features. For InternVL, it uses the visual tokens returned by the model's
feature-extraction module after pixel shuffle and MLP projection.

\paragraph{Optimization.}
All downstream-agnostic attacks use an $\ell_\infty$ perturbation budget of
$4/255$ and 100 optimization steps. CIRA starts from the clean image and uses
projected sign-gradient ascent with step size $1/255$, $\lambda=0.8$, and
candidate budget range $[K_{\min},K_{\max}]=[32,192]$. The remaining
hyperparameters of VEAttack \citep{mei2026veattack} and CAGE
\citep{zhang2026adversarial} follow their released settings; CAGE uses its released
budget interval $[16,192]$. CAA$^{\dagger}$ \citep{zhang2026less} optimizes a question-conditioned
objective at language-model layer 2 with $\epsilon=4/255$ and 100 steps.
One adversarial image is generated per image--question pair and then
evaluated across compressors and budgets without re-optimization.

\paragraph{Evaluation protocol.}
Each downstream-agnostic method generates one adversarial image per clean
input. After optimization, the image is fixed and evaluated across all
corresponding questions, compressors, and budgets. Clean eligibility is
determined separately for each compressor--budget setting using
$\mathcal{S}_{K}$ in \eqref{eq:clean_eligible_set}, while Full ASR is computed
over $\mathcal{S}_{\mathrm{full}}$ defined in
Section~\ref{sec:evaluation_metrics}.

%% file: appendix/diagnostics.tex
\section{Controlled Retained-Set Allocation Diagnostics}
\label{sec:appendix_motivation}

The diagnostic in Section~\ref{sec:motivation} isolates retained-set allocation
through counterfactual exchanges at fixed first-observed CSF states.
Answer-aware scores are used only for post-hoc exchange selection after the
failure state is fixed and are not part of \textsc{CIRA} optimization.

\subsection{Diagnostic Cohort and Fixed Failure State}
\label{sec:appendix_first_csf}

\paragraph{Diagnostic setting.}
We use LLaVA--VisionZip on POPE, TextVQA, and MME at
$K\in\{32,64,128,192\}$. VisionZip retains dominant patch tokens individually
and aggregates the remainder into contextual tokens. We call the individually
retained patches \emph{direct tokens}; their capacity $D_K$ is smaller than the
total compressed budget $K$.

\paragraph{Common clean cohort.}
Let $j$ index an image--question observation, comprising image $x_j$, its
associated question, and reference answer; index $i$ is reserved for visual
tokens. We retain only observations that are correct under full-token inference and under
every evaluated compressed setting:
\begin{equation}
c_j(x_j)=1,
\qquad
c_{j,K}(x_j)=1,
\quad
\forall K\in\{192,128,64,32\},
\end{equation}
where $c_j$ and $c_{j,K}$ denote full-token and compressed correctness for
observation $j$. Every trajectory therefore begins from a common state without
pre-existing compression errors in any evaluated path.

\paragraph{Attack trajectory.}
We generate clean-initialized feature-objective VEAttack \citep{mei2026veattack}
trajectories with
$\epsilon=4/255$, step size $1/255$, and at most $100$ steps, evaluating
predictions every $10$ steps. For each observation and setting, we select the
earliest evaluated checkpoint satisfying
\begin{equation}
c_j(x_j^{\mathrm{adv}})=1,
\qquad
c_{j,K}(x_j^{\mathrm{adv}})=0.
\end{equation}
This \emph{first-observed CSF} supplies the fixed state for all subsequent
counterfactuals.

\paragraph{Balanced diagnostic cohort.}
Within each comparison, eligible images are ordered using a fixed sampling
order and truncated to the smallest available image count across cells. All
eligible questions associated with the retained images are then included.

\subsection{Controlled Counterfactual Reallocation}
\label{sec:appendix_reallocation}

At the fixed adversarial state, exchange identities are selected using
retrospective answer support computed from the adversarial representations.
For adversarial token $i$, we apply a multiplicative gate
$\widetilde h_i=g_i h_i^{\mathrm{adv}}$ and define
\begin{equation}
a_i
=
-
\left.
\frac{\partial
\mathcal L_{\mathrm{NLL}}
(y\mid \widetilde H^{\mathrm{adv}},q)}
{\partial g_i}
\right|_{g_i=1}.
\label{eq:appendix_evidence_score}
\end{equation}
Larger $a_i$ indicates that token $i$ provides stronger support for the reference answer.

Let $R_K$ be the direct-token set and $O_K$ its complement. We rank $O_K$ in
descending and $R_K$ in ascending order of $a_i$. For exchange size $r$,
\begin{equation}
R_{K,r}^{\mathrm{guide}}
=
\left(R_K\setminus L_r\right)\cup H_r,
\end{equation}
where $H_r$ contains the $r$ highest-support omitted tokens and $L_r$ the $r$
lowest-support retained tokens.

We evaluate the nominal schedule
$\rho\in\{0,5\%,7.5\%,10\%,15\%\}$, with
\begin{equation}
r_K(\rho)
=
\operatorname{round}(\rho D_K),
\end{equation}
where $D_K$ is the number of direct slots. For
$\rho=(0,5\%,7.5\%,10\%,15\%)$, this gives
$r_{32}=(0,1,2,3,4)$, $r_{64}=(0,3,4,5,8)$,
$r_{128}=(0,5,8,11,16)$, and $r_{192}=(0,8,12,16,24)$.

The matched-random control independently permutes the retained and omitted
pools. Each observation uses 40 nested paths, with larger exchanges extending
smaller ones. VisionZip merging and aggregation are recomputed after every
exchange.

\subsection{Retained-Set Allocation Sensitivity and Cumulative Recovery}
\label{sec:appendix_repairability_metric}

Let $Y_{jKz,p}(\rho)\in\{0,1\}$ indicate correctness for observation $j$ under
condition $z$, path $p$, and ratio $\rho$. Guided reallocation has one path;
the random control has $P=40$ nested paths. Because recovery need not persist
under a larger exchange, curves report cumulative recovery:
\begin{equation}
C_{jKz,p}(\rho)
=
\max_{\rho'\le\rho}
Y_{jKz,p}(\rho').
\label{eq:cumulative_repairability}
\end{equation}
Thus, $C_{jKz,p}(\rho)$ asks whether a CSF is corrected at any schedule point
up to the displayed nominal ratio.

\paragraph{Estimand and aggregation.}
Random paths are averaged within each observation before cell aggregation:
\begin{equation}
\overline{C}_{jK,\mathrm{rand}}(\rho)
=
\frac{1}{P}\sum_{p=1}^{P} C_{jK,\mathrm{rand},p}(\rho),
\label{eq:appendix_random_path_collapse}
\end{equation}
and we set $\overline{C}_{jK,\mathrm{guide}}(\rho)=
C_{jK,\mathrm{guide},1}(\rho)$. Let $\mathcal{G}$ denote the 12
dataset--budget cells and $\mathcal{J}_{dK}$ their analyzed image--question
observations. The reported equal-cell average is
\begin{equation}
\widehat{\mu}_{z}(\rho)
=
\frac{1}{|\mathcal{G}|}
\sum_{(d,K)\in\mathcal{G}}
\left[
\frac{1}{|\mathcal{J}_{dK}|}
\sum_{j\in\mathcal{J}_{dK}}
\overline{C}_{jK,z}(\rho)
\right].
\label{eq:appendix_repairability_average}
\end{equation}
The inner mean averages over image--question observations within each cell,
while the outer mean assigns equal weight to every dataset--budget cell. The paired contrast is $\widehat{\Delta}(\rho)=
\widehat{\mu}_{\mathrm{guide}}(\rho)-
\widehat{\mu}_{\mathrm{rand}}(\rho)$.

\input{tables/guided_recovery}

The average guided--random advantage ranges from $22.5$ to $24.1$ pp
across all nonzero exchange ratios.

\subsection{Factorial Decomposition of Recovery}
\label{sec:appendix_factorial}

At $\rho^{*}=7.5\%$, we separate which omitted tokens enter
from which retained tokens leave. Incoming tokens are high-support evidence or
random omissions; outgoing tokens are low-support retained tokens or random
ones:
\[
\begin{array}{c|cc}
 & \text{Random out} & \text{Low-support out} \\
\hline
\text{Random in}   & RR & RL \\
\text{Evidence in} & ER & EL
\end{array}
\]
Let $Q_{jK,z}\in[0,1]$ denote exact recovery under arm $z$, averaged over
random paths where applicable. We compute the two main effects and their
interaction per observation before equal-cell aggregation:
\begin{align}
e_{jK}
&=
\frac{
(Q_{jK,ER}-Q_{jK,RR})+
(Q_{jK,EL}-Q_{jK,RL})
}{2},\\
u_{jK}
&=
\frac{
(Q_{jK,RL}-Q_{jK,RR})+
(Q_{jK,EL}-Q_{jK,ER})
}{2},\\
\eta_{jK}
&=
Q_{jK,EL}-Q_{jK,ER}-Q_{jK,RL}+Q_{jK,RR}.
\label{eq:appendix_observation_factorial_effects}
\end{align}
The four arm rates and observation-level effects are aggregated with
\eqref{eq:appendix_repairability_average};
Table~\ref{tab:appendix_detailed_factorial} reports the resulting point
estimates by budget.

\input{tables/factorial_recovery}

Evidence restoration is positive at every budget, while the removal effect
increases with $K$. Interaction estimates are negative and smaller in magnitude
than either main effect at every budget.

\subsection{Representation-Drift Moderation}
\label{sec:appendix_representation_shift}

For a restored evidence token $i$, let $h_i^{\mathrm{clean}}$ and
$h_i^{\mathrm{adv}}$ be its clean and adversarial encoder representations. We
define
\begin{equation}
d_i^{\mathrm{diag}}
=
\frac{
1-\cos
\left(
h_i^{\mathrm{clean}},
h_i^{\mathrm{adv}}
\right)
}{2}.
\end{equation}
Let $E_{jK}$ denote the high-support omitted tokens restored for observation
$j$ at budget $K$ under the $\rho^{*}=7.5\%$ intervention. We define their mean
representation drift as
\begin{equation}
D_{jK}
=
\frac{1}{|E_{jK}|}
\sum_{i\in E_{jK}} d_i^{\mathrm{diag}}.
\label{eq:evidence_shift}
\end{equation}
Clean representations are used only in this post-hoc diagnostic. Let
$\overline{D}_{dK}$ and $\overline{e}_{dK}$ be within-cell means. We estimate
the common within-cell association between drift and the evidence-restoration
effect $e_{jK}$ from \eqref{eq:appendix_observation_factorial_effects}:
\begin{equation}
\widehat{\beta}
=
\frac{
\sum_{d,K}\sum_{j\in\mathcal{J}_{dK}}
\bigl(D_{jK}-\overline{D}_{dK}\bigr)
\bigl(e_{jK}-\overline{e}_{dK}\bigr)
}{
\sum_{d,K}\sum_{j\in\mathcal{J}_{dK}}
(D_{jK}-\overline{D}_{dK})^2
}.
\label{eq:appendix_damage_fixed_effect_slope}
\end{equation}
Thus, $\widehat{\beta}$ uses only within-cell variation. Drift tertiles provide
a grouped summary, and the continuous slope summarizes the corresponding
within-cell association.

\input{tables/representation_drift}

The estimated within-cell association is $-8.1$ pp per $0.1$ increase in
$D_{jK}$. Budget-specific tertiles are not uniformly monotone
and are therefore interpreted descriptively. The retained-set intervention
establishes that changing token allocation can causally alter correctness within
this fixed cohort, whereas the drift--recovery analysis supports an association
rather than causal mediation.

%% file: tables/guided_recovery.tex
\begin{table*}[htbp]
\centering
\scriptsize
\caption{Cumulative guided (G) and matched-random (R) recovery (\%) by compression
budget and exchange ratio, with $\Delta=G-R$ reported in percentage points and
Avg. denoting the equal-cell average.}
\label{tab:appendix_detailed_repairability}
\vspace{2pt}
\begin{adjustbox}{max width=\textwidth}
\renewcommand{\arraystretch}{1.12}
\setlength{\tabcolsep}{2.7pt}
\begin{tabular}{
c
!{\color{black}\vrule width 0.45pt}
ccc
!{\color{black}\vrule width 0.45pt}
ccc
!{\color{black}\vrule width 0.45pt}
ccc
!{\color{black}\vrule width 0.45pt}
ccc
}
\Xhline{2pt}
\rowcolor{headercolor}
\textbf{$K$}
& \multicolumn{3}{c!{\color{black}\vrule width 0.45pt}}{\textbf{$\rho=5\%$}}
& \multicolumn{3}{c!{\color{black}\vrule width 0.45pt}}{\textbf{$\rho=7.5\%$}}
& \multicolumn{3}{c!{\color{black}\vrule width 0.45pt}}{\textbf{$\rho=10\%$}}
& \multicolumn{3}{c}{\textbf{$\rho=15\%$}} \\
\rowcolor{headercolor}
{}
& \textbf{G} & \textbf{R} & $\boldsymbol{\Delta}$
& \textbf{G} & \textbf{R} & $\boldsymbol{\Delta}$
& \textbf{G} & \textbf{R} & $\boldsymbol{\Delta}$
& \textbf{G} & \textbf{R} & $\boldsymbol{\Delta}$ \\
\Xhline{2pt}
32
& 29.1 & 12.3 & $+16.8$ & 40.6 & 19.3 & $+21.4$
& 45.5 & 23.6 & $+21.9$ & 51.1 & 27.5 & $+23.6$ \\
64
& 39.1 & 16.5 & $+22.6$ & 43.4 & 21.8 & $+21.6$
& 44.1 & 25.9 & $+18.3$ & 49.7 & 32.5 & $+17.2$ \\
128
& 35.8 & 13.8 & $+22.0$ & 45.8 & 20.9 & $+24.9$
& 49.4 & 25.4 & $+24.0$ & 54.9 & 30.8 & $+24.1$ \\
192
& 44.4 & 15.6 & $+28.8$ & 50.8 & 22.3 & $+28.4$
& 53.6 & 27.0 & $+26.6$ & 57.9 & 33.0 & $+24.9$ \\
\rowcolor{gray!12}
\textbf{Avg.}
& \textbf{37.1} & \textbf{14.5} & $\mathbf{+22.6}$
& \textbf{45.2} & \textbf{21.1} & $\mathbf{+24.1}$
& \textbf{48.2} & \textbf{25.5} & $\mathbf{+22.7}$
& \textbf{53.4} & \textbf{31.0} & $\mathbf{+22.5}$ \\
\Xhline{2pt}
\end{tabular}
\end{adjustbox}
\end{table*}

%% file: tables/factorial_recovery.tex
\begin{table*}[htbp]
\centering
\scriptsize
\caption{Exact recovery and factorial effects at $\rho^{*}=7.5\%$ by
compression budget, with Avg. denoting the equal-cell average.}
\label{tab:appendix_detailed_factorial}
\vspace{2pt}
\begin{adjustbox}{max width=\textwidth}
\renewcommand{\arraystretch}{1.12}
\setlength{\tabcolsep}{3.4pt}
\begin{tabular}{
c
!{\color{black}\vrule width 0.45pt}
cccc
!{\color{black}\vrule width 0.45pt}
ccc
}
\Xhline{2pt}
\rowcolor{headercolor}
\textbf{$K$}
& \multicolumn{4}{c!{\color{black}\vrule width 0.45pt}}{\textbf{Exact Recovery (\%)}}
& \multicolumn{3}{c}{\textbf{Factorial Effect (pp)}} \\
\rowcolor{headercolor}
{}
& \textbf{RR} & \textbf{ER} & \textbf{RL} & \textbf{EL}
& \textbf{Evidence Restoration}
& \textbf{Low-Support Removal}
& \textbf{Interaction} \\
\Xhline{2pt}
32
& 16.1 & 34.9 & 20.1 & 38.6
& $+18.6$
& $+3.8$
& $-0.3$ \\
64
& 18.4 & 34.0 & 25.4 & 38.5
& $+14.4$
& $+5.7$
& $-2.4$ \\
128
& 17.2 & 33.5 & 31.9 & 44.4
& $+14.4$
& $+12.8$
& $-3.9$ \\
192
& 17.9 & 31.1 & 34.5 & 46.4
& $+12.6$
& $+15.9$
& $-1.3$ \\
\rowcolor{gray!12}
\textbf{Avg.}
& \textbf{17.4} & \textbf{33.4} & \textbf{28.0} & \textbf{42.0}
& $\mathbf{+15.0}$
& $\mathbf{+9.6}$
& $-2.0$ \\
\Xhline{2pt}
\end{tabular}
\end{adjustbox}
\end{table*}

%% file: tables/representation_drift.tex
\begin{table*}[htbp]
\centering
\scriptsize
\caption{Evidence-restoration effects across within-cell representation-drift
tertiles and compression budgets, with mean drift defined as $100\times$
half-cosine distance and Gap as the lower-minus-upper effect.}
\label{tab:appendix_detailed_shift_moderation}
\vspace{2pt}
\begin{adjustbox}{max width=\textwidth}
\renewcommand{\arraystretch}{1.12}
\setlength{\tabcolsep}{2.7pt}
\begin{tabular}{
c
!{\color{black}\vrule width 0.45pt}
cc
!{\color{black}\vrule width 0.45pt}
cc
!{\color{black}\vrule width 0.45pt}
cc
!{\color{black}\vrule width 0.45pt}
c
}
\Xhline{2pt}
\rowcolor{headercolor}
\textbf{$K$}
& \multicolumn{2}{c!{\color{black}\vrule width 0.45pt}}{\textbf{Lower}}
& \multicolumn{2}{c!{\color{black}\vrule width 0.45pt}}{\textbf{Middle}}
& \multicolumn{2}{c!{\color{black}\vrule width 0.45pt}}{\textbf{Upper}}
& \textbf{Gap (pp)} \\
\rowcolor{headercolor}
{}
& \textbf{Drift (\%)} & \textbf{Effect (pp)}
& \textbf{Drift (\%)} & \textbf{Effect (pp)}
& \textbf{Drift (\%)} & \textbf{Effect (pp)}
& {} \\
\Xhline{2pt}
32
& 26.1 & $+28.5$
& 38.0 & $+20.8$
& 45.2 & $+6.1$
& $+22.3$ \\
64
& 31.0 & $+18.2$
& 39.3 & $+7.0$
& 45.5 & $+18.0$
& $+0.2$ \\
128
& 34.0 & $+18.9$
& 40.4 & $+12.7$
& 45.1 & $+11.5$
& $+7.4$ \\
192
& 34.4 & $+16.6$
& 40.2 & $+15.3$
& 45.1 & $+5.4$
& $+11.2$ \\
\rowcolor{gray!12}
\textbf{Avg.}
& \textbf{31.4} & $\mathbf{+20.5}$
& \textbf{39.5} & $\mathbf{+14.0}$
& \textbf{45.2} & $\mathbf{+10.3}$
& $\mathbf{+10.3}$ \\
\Xhline{2pt}
\end{tabular}
\end{adjustbox}
\end{table*}

%% file: appendix/sensitivity.tex
\section{Sensitivity Analyses}
\label{sec:appendix_sensitivity}

    We test whether CIRA's selective operating point depends on individual design
    or optimization choices. Unless stated otherwise, each analysis varies one
    choice on LLaVA-v1.5-7B, POPE, and VisionZip while retaining the evaluation
    definitions in Section~\ref{sec:evaluation_metrics}.

\subsection{Preservation Weight}
\label{sec:appendix_lambda}

The relative loss weight $\lambda$ in \eqref{eq:joint_cira_loss} controls the
tradeoff between failure induction and full-token preservation. We summarize
this operating tradeoff by Selective Gap (Avg.\ CSFR minus Full ASR), used only
as a configuration score because the two metrics have different conditioning
sets. Table~\ref{tab:lambda_sensitivity} shows that increasing $\lambda$ reduces
Full ASR while retaining substantial CSFR. Selective Gap remains stable for
$\lambda\in[0.6,1.0]$, with $\lambda=0.8$ attaining the highest observed value.
We therefore use $\lambda=0.8$ throughout the evaluation.

\input{tables/lambda_sensitivity}

\subsection{Candidate Compression-Budget Interval}
\label{sec:appendix_k_interval}

The admissible interval $[K_{\min},K_{\max}]$ determines the budget-marginal
displacement weights in \eqref{eq:hidden_weight}.
Table~\ref{tab:k_interval_sensitivity} varies this interval while keeping
the evaluation budgets fixed at $K\in\{192,128,64,32\}$. Selectivity varies modestly across the six candidate intervals. We use $[32,192]$ as the default because it matches the evaluation range; its Selective Gap is within $0.28$ pp of the best observed value.

\input{tables/budget_interval_sensitivity}

\subsection{Perturbation Budget and Optimization Steps}
\label{sec:appendix_attack_parameters}

\begin{figure*}[htbp]
    \centering
    \includegraphics[width=1.0\textwidth]{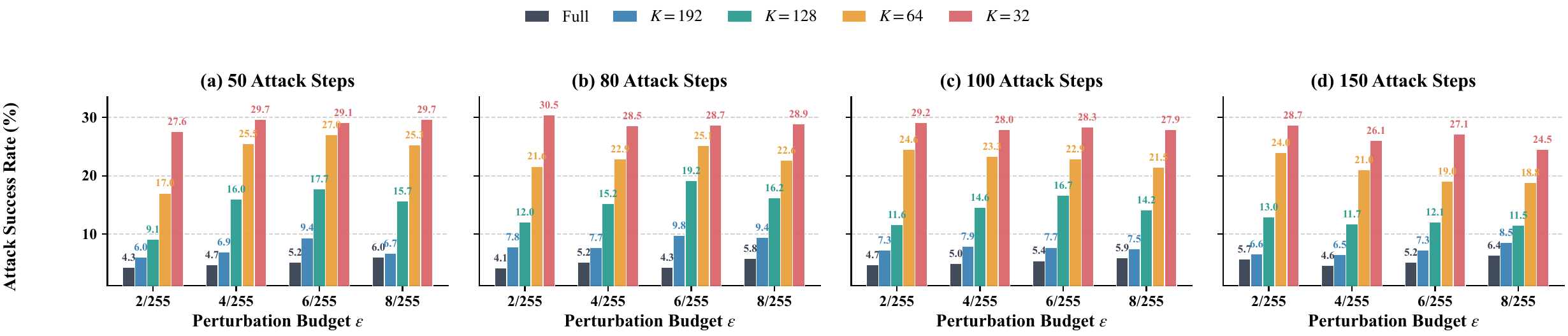}
    \caption{Sensitivity of Full ASR and CSFR to the perturbation budget and
    number of optimization steps on POPE.}
    \label{fig:attack_parameter_sensitivity}
\end{figure*}

Figure~\ref{fig:attack_parameter_sensitivity} varies
$\epsilon\in\{2,4,6,8\}/255$ and the number of optimization steps in
$\{50,80,100,150\}$; the step size is set to $\epsilon/4$. Across all 16 configurations, Full ASR remains between $4.14\%$ and $6.38\%$,
whereas Avg.\ CSFR ranges from $14.93\%$ to $20.82\%$. Thus, substantial
compression-specific failure rates persist while full-token degradation remains
limited, and the default $\epsilon=4/255$, 100-step setting lies within a stable
selective region rather than at an isolated optimum.

\subsection{Scoring-Layer Configuration Study}
\label{sec:appendix_layer_selection}

The primary score in \eqref{eq:token_importance} averages class-to-patch
attention over the scoring-layer configuration
$\mathcal{L}_{\mathrm{score}}$. We compare 14 single-layer, contiguous, and spaced multi-layer configurations using 100-example design subsets from POPE, TextVQA, and MME, evaluated with VisionZip, PruMerge, and VisPruner. We select the configuration with the highest mean Selective Gap across these nine dataset--compressor environments and use it throughout the reported evaluation.

\begin{figure*}[htbp]
    \centering
    \includegraphics[width=1.0\textwidth]{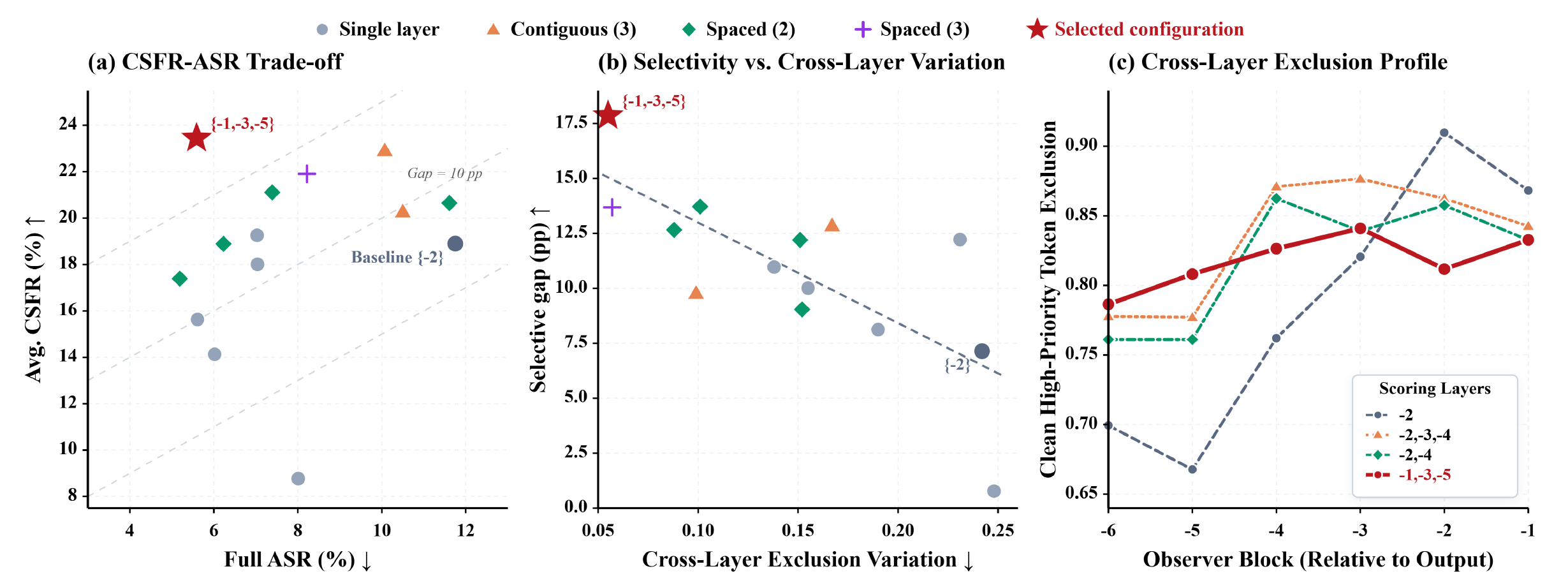}
    \caption{Selectivity and cross-layer exclusion stability across scoring-layer
    configurations.}
    \label{fig:layer_selection_sensitivity}
\end{figure*}

As shown in Figure~\ref{fig:layer_selection_sensitivity}, the selected spaced
late-layer triplet $\{-1,-3,-5\}$ achieves $23.45\%$ Avg.\ CSFR with $5.59\%$ Full ASR, corresponding to a $17.86$ pp Selective Gap. Its exclusion rate for clean high-priority tokens also varies less across observer layers than that of the single-layer $\{-2\}$ configuration (range $0.055$ versus $0.242$). The negative association between cross-layer variation and selectivity ($\rho_s=-0.70$) indicates that configurations with more stable exclusion behavior across depth tend to exhibit higher selectivity. This association is descriptive rather than causal.

%% file: tables/lambda_sensitivity.tex
\begin{table*}[htbp]
\centering
\footnotesize

\caption{Sensitivity of Full ASR, CSFR, and Selective Gap to the preservation
weight $\lambda$ on POPE. Best values in each row are shown in \textbf{bold}.}
\label{tab:lambda_sensitivity}

\vspace{1pt}

\renewcommand{\arraystretch}{1.15}
\setlength{\tabcolsep}{7.0pt}

\begin{adjustbox}{max width=0.78\textwidth}
\begin{tabular}{
l
!{\color{black}\vrule width 0.45pt}
ccccc
}

\Xhline{2pt}

\rowcolor{headercolor}
\textbf{Metric}
& $\boldsymbol{\lambda=0.2}$
& $\boldsymbol{\lambda=0.4}$
& $\boldsymbol{\lambda=0.6}$
& $\boldsymbol{\lambda=0.8}$
& $\boldsymbol{\lambda=1.0}$
\\

\Xhline{2pt}

\textbf{Full ASR}\textcolor{blue}{$\downarrow$}
& 9.57
& 6.03
& 5.79
& 4.96
& \textbf{4.61}
\\

\Xhline{0.45pt}

\textbf{CSFR@192}\textcolor{red}{$\uparrow$}
& 8.59
& 9.08
& \textbf{9.20}
& 7.61
& 6.50
\\

\textbf{CSFR@128}\textcolor{red}{$\uparrow$}
& \textbf{15.67}
& 15.30
& 15.55
& 14.05
& 13.68
\\

\textbf{CSFR@64}\textcolor{red}{$\uparrow$}
& 22.07
& 22.07
& \textbf{22.47}
& \textbf{22.47}
& 21.94
\\

\textbf{CSFR@32}\textcolor{red}{$\uparrow$}
& \textbf{27.56}
& 27.26
& 25.93
& 26.96
& 27.41
\\

\textbf{Avg. CSFR}\textcolor{red}{$\uparrow$}
& \textbf{18.47}
& 18.43
& 18.29
& 17.77
& 17.38
\\

\Xhline{0.45pt}

\rowcolor{gray!12}
\textbf{Selective Gap}\textcolor{red}{$\uparrow$}
& 8.90
& 12.40
& 12.50
& \textbf{12.81}
& 12.77
\\

\Xhline{2pt}

\end{tabular}
\end{adjustbox}

\vspace{-3pt}
\end{table*}

%% file: tables/budget_interval_sensitivity.tex
\begin{table*}[htbp]
\centering
\footnotesize

\caption{Sensitivity of Full ASR, CSFR, and Selective Gap to the candidate
compression-budget interval on POPE. Best values in each row are shown in \textbf{bold}.}
\label{tab:k_interval_sensitivity}

\vspace{1pt}

\renewcommand{\arraystretch}{1.15}
\setlength{\tabcolsep}{5.5pt}

\begin{adjustbox}{max width=0.95\textwidth}
\begin{tabular}{
l
!{\color{black}\vrule width 0.45pt}
c
!{\color{black}\vrule width 0.45pt}
cc
!{\color{black}\vrule width 0.45pt}
ccc
}

\Xhline{2pt}

\rowcolor{headercolor}
\textbf{Metric}
& \textbf{Default}
& \multicolumn{2}{
c!{\color{black}\vrule width 0.45pt}
}{\textbf{Vary $K_{\min}$}}
& \multicolumn{3}{c}{\textbf{Vary $K_{\max}$}}
\\

\rowcolor{headercolor}
{}
& $\boldsymbol{[32,192]}$
& $\boldsymbol{[16,192]}$
& $\boldsymbol{[64,192]}$
& $\boldsymbol{[32,64]}$
& $\boldsymbol{[32,128]}$
& $\boldsymbol{[32,384]}$
\\

\Xhline{2pt}

\textbf{Full ASR}\textcolor{blue}{$\downarrow$}
& 4.96
& \textbf{4.62}
& 5.56
& 5.68
& 4.97
& 5.33
\\

\Xhline{0.45pt}

\textbf{CSFR@192}\textcolor{red}{$\uparrow$}
& 7.61
& 7.38
& 6.40
& \textbf{8.86}
& 7.63
& 6.52
\\

\textbf{CSFR@128}\textcolor{red}{$\uparrow$}
& 14.05
& \textbf{14.86}
& 12.86
& 13.23
& 14.73
& 12.86
\\

\textbf{CSFR@64}\textcolor{red}{$\uparrow$}
& 22.47
& 21.14
& 21.68
& \textbf{23.01}
& 19.95
& 22.61
\\

\textbf{CSFR@32}\textcolor{red}{$\uparrow$}
& 26.96
& 27.43
& 28.17
& \textbf{29.20}
& 26.55
& 25.22
\\

\textbf{Avg. CSFR}\textcolor{red}{$\uparrow$}
& 17.77
& 17.70
& 17.28
& \textbf{18.57}
& 17.21
& 16.80
\\

\Xhline{0.45pt}

\rowcolor{gray!12}
\textbf{Selective Gap}\textcolor{red}{$\uparrow$}
& 12.81
& \textbf{13.09}
& 11.71
& 12.89
& 12.24
& 11.48
\\

\Xhline{2pt}

\end{tabular}
\end{adjustbox}

\vspace{-3pt}
\end{table*}

%% file: appendix/cross_model.tex
\section{Cross-Model Scope}
\label{sec:appendix_cross_model}

The main experiments establish CIRA across multiple compressors and budgets on
LLaVA. We next test whether paired selectivity persists when the vision encoder,
multimodal interface, and language model change together, using
Qwen3-VL-8B-Instruct \citep{bai2025qwen3} and InternVL3.5-8B
\citep{wang2025internvl3}.

Here $K$ denotes the number of compressed visual tokens passed to subsequent
computation in each model's native interface. For compact reporting, we use
$K_1/64/32$, where $K_1=96$ for Qwen3-VL and $K_1=128$ for InternVL.
Avg.\ CSFR is the arithmetic mean over the three budgets within each model
family. Table~\ref{tab:qwen_internvl_compact_results} reports the corresponding
CSFR and Full ASR results, while clean accuracy is provided in
Table~\ref{tab:clean_utility}.

\input{tables/cross_model_results}

Across both additional model families, CIRA continues to induce
compression-specific failures while keeping Full ASR substantially below
those of VEAttack and CAGE. The pattern is strongest on TextVQA, where CIRA achieves
the highest Avg.\ CSFR among downstream-agnostic attacks for every
compressor on both model families. Results on
POPE and MME are more heterogeneous across compressors, but compression-specific
failure induction remains observable under target-encoder-only access.
Together, these results show that CIRA's compression-selective behavior extends
beyond LLaVA to distinct vision encoders and native visual-token interfaces.

%% file: tables/cross_model_results.tex
\begin{table*}[htbp]
\centering
\scriptsize

\caption{CSFR and Full ASR on Qwen3-VL-8B-Instruct and InternVL3.5-8B.
Compressor entries report CSFR at $K_1/64/32$ followed by their average.
Bold and underlined values denote the best and second-best downstream-agnostic
results, respectively; CAA$^{\dagger}$ is shown as a stronger-access reference
and excluded from these rankings.}
\label{tab:qwen_internvl_compact_results}

\begin{adjustbox}{max width=\textwidth}
\renewcommand{\arraystretch}{1.10}
\setlength{\tabcolsep}{3.2pt}

\begin{tabular}{
>{\centering\arraybackslash}m{1.25cm}
>{\centering\arraybackslash}m{1.55cm}
!{\color{black}\vrule width 0.45pt}
c
!{\color{black}\vrule width 0.45pt}
cccc
}

\Xhline{2pt}

\rowcolor{headercolor}
\textbf{Dataset}
& \textbf{Attack}
& \shortstack[c]{
    \textbf{Full ASR}\tabularnewline[-1pt]
    {\scriptsize(\%)\textcolor{blue}{$\downarrow$}}
  }
& \shortstack[c]{
    \textbf{VisionZip}\tabularnewline[-1pt]
    {\scriptsize$K_1/64/32/\mathrm{Avg.}$\textcolor{red}{$\uparrow$}}
  }
& \shortstack[c]{
    \textbf{VisPruner}\tabularnewline[-1pt]
    {\scriptsize$K_1/64/32/\mathrm{Avg.}$\textcolor{red}{$\uparrow$}}
  }
& \shortstack[c]{
    \textbf{PruMerge}\tabularnewline[-1pt]
    {\scriptsize$K_1/64/32/\mathrm{Avg.}$\textcolor{red}{$\uparrow$}}
  }
& \shortstack[c]{
    \textbf{FastV}\tabularnewline[-1pt]
    {\scriptsize$K_1/64/32/\mathrm{Avg.}$\textcolor{red}{$\uparrow$}}
  }
\\

\Xhline{2pt}

% Qwen3-VL-8B-Instruct
\multicolumn{7}{>{\columncolor{gray!20}}c}{
\textbf{Qwen3-VL-8B-Instruct}
\quad ($K_1=96$)
}
\\
\hline

% Qwen: POPE
\multirow{4}{*}{\textbf{POPE}}
& VEAttack
& \Second{45.93}
& \CSFRTuple{4.63}{5.83}{10.06}{6.84}
& \CSFRTuple{2.15}{2.78}{4.67}{3.20}
& \CSFRTuple{\Second{3.91}}{\Second{5.01}}{7.49}{\Second{5.47}}
& \CSFRTuple{2.32}{3.95}{4.60}{3.62}
\\

& CAGE
& 46.47
& \CSFRTuple
  {\Second{7.13}}
  {\Second{8.13}}
  {\Second{10.19}}
  {\Second{8.48}}
& \CSFRTuple
  {\Second{4.65}}
  {\Second{5.68}}
  {\Second{8.32}}
  {\Second{6.22}}
& \CSFRTuple
  {\Best{7.23}}
  {\Best{8.23}}
  {\Best{8.95}}
  {\Best{8.14}}
& \CSFRTuple
  {\Second{4.65}}
  {\Second{6.25}}
  {\Second{11.49}}
  {\Second{7.46}}
\\

& CAA$^{\dagger}$
& 4.40
& \CSFRTuple{4.03}{5.82}{11.31}{7.05}
& \CSFRTuple{2.61}{4.86}{8.70}{5.39}
& \CSFRTuple{2.47}{3.33}{7.06}{4.29}
& \CSFRTuple{8.44}{11.61}{14.80}{11.61}
\\

& \textbf{CIRA}
& \Best{10.66}
& \CSFRTuple
  {\Best{7.59}}
  {\Best{9.82}}
  {\Best{14.14}}
  {\Best{10.52}}
& \CSFRTuple
  {\Best{4.98}}
  {\Best{6.56}}
  {\Best{11.22}}
  {\Best{7.59}}
& \CSFRTuple
  {3.65}
  {4.39}
  {\Second{7.94}}
  {5.33}
& \CSFRTuple
  {\Best{8.56}}
  {\Best{10.08}}
  {\Best{13.07}}
  {\Best{10.57}}
\\

\Xhline{1.1pt}

% Qwen: TextVQA
\multirow{4}{*}{\textbf{TextVQA}}
& VEAttack
& 80.14
& \CSFRTuple{10.11}{10.03}{3.58}{7.91}
& \CSFRTuple{10.14}{7.75}{1.98}{6.62}
& \CSFRTuple{10.22}{7.87}{3.34}{7.14}
& \CSFRTuple{3.54}{3.58}{3.65}{3.59}
\\

& CAGE
& \Second{66.03}
& \CSFRTuple
  {\Second{23.96}}
  {\Second{22.56}}
  {\Second{15.22}}
  {\Second{20.58}}
& \CSFRTuple
  {\Second{19.59}}
  {\Second{16.71}}
  {\Second{10.62}}
  {\Second{15.64}}
& \CSFRTuple
  {\Second{21.54}}
  {\Second{18.12}}
  {\Second{16.56}}
  {\Second{18.74}}
& \CSFRTuple
  {\Second{12.38}}
  {\Second{11.00}}
  {\Second{12.41}}
  {\Second{11.93}}
\\

& CAA$^{\dagger}$
& 8.24
& \CSFRTuple{46.37}{42.86}{35.22}{41.48}
& \CSFRTuple{29.28}{31.23}{32.84}{31.12}
& \CSFRTuple{31.69}{30.19}{31.91}{31.27}
& \CSFRTuple{43.81}{42.20}{38.69}{41.57}
\\

& \textbf{CIRA}
& \Best{16.37}
& \CSFRTuple
  {\Best{51.65}}
  {\Best{52.13}}
  {\Best{45.37}}
  {\Best{49.72}}
& \CSFRTuple
  {\Best{30.63}}
  {\Best{28.81}}
  {\Best{34.57}}
  {\Best{31.34}}
& \CSFRTuple
  {\Best{34.88}}
  {\Best{32.13}}
  {\Best{35.20}}
  {\Best{34.07}}
& \CSFRTuple
  {\Best{38.31}}
  {\Best{36.57}}
  {\Best{34.31}}
  {\Best{36.40}}
\\

\Xhline{1.1pt}

% Qwen: MME
\multirow{4}{*}{\textbf{MME}}
& VEAttack
& 44.11
& \CSFRTuple{7.14}{6.59}{9.29}{7.67}
& \CSFRTuple{5.95}{6.51}{6.29}{6.25}
& \CSFRTuple
  {\Second{7.60}}
  {\Second{8.31}}
  {\Second{9.69}}
  {\Second{8.54}}
& \CSFRTuple{2.58}{3.84}{6.58}{4.33}
\\

& CAGE
& \Second{41.60}
& \CSFRTuple
  {\Best{11.67}}
  {\Second{7.78}}
  {\Second{10.19}}
  {\Best{9.88}}
& \CSFRTuple
  {\Best{11.41}}
  {\Best{7.83}}
  {\Second{9.37}}
  {\Best{9.54}}
& \CSFRTuple
  {\Best{12.47}}
  {\Best{11.87}}
  {\Best{15.09}}
  {\Best{13.14}}
& \CSFRTuple
  {\Second{4.42}}
  {\Second{6.64}}
  {\Second{8.53}}
  {\Second{6.53}}
\\

& CAA$^{\dagger}$
& 4.76
& \CSFRTuple{8.27}{9.96}{12.35}{10.19}
& \CSFRTuple{7.37}{7.72}{10.34}{8.48}
& \CSFRTuple{9.71}{9.72}{12.08}{10.51}
& \CSFRTuple{9.16}{11.71}{15.66}{12.18}
\\

& \textbf{CIRA}
& \Best{12.43}
& \CSFRTuple
  {\Second{8.34}}
  {\Best{7.79}}
  {\Best{12.79}}
  {\Second{9.64}}
& \CSFRTuple
  {\Second{8.82}}
  {\Second{7.37}}
  {\Best{9.91}}
  {\Second{8.70}}
& \CSFRTuple{7.14}{5.26}{7.05}{6.49}
& \CSFRTuple
  {\Best{5.28}}
  {\Best{6.77}}
  {\Best{10.21}}
  {\Best{7.42}}
\\

\Xhline{1.5pt}

% InternVL3.5-8B
\multicolumn{7}{>{\columncolor{gray!20}}c}{
\textbf{InternVL3.5-8B}
\quad ($K_1=128$)
}
\\
\hline

% InternVL: POPE
\multirow{4}{*}{\textbf{POPE}}
& VEAttack
& 51.46
& \CSFRTuple
  {\Second{6.01}}
  {\Second{8.17}}
  {10.35}
  {\Second{8.18}}
& \CSFRTuple
  {\Best{4.20}}
  {\Best{7.44}}
  {7.91}
  {\Best{6.52}}
& \CSFRTuple
  {\Best{5.28}}
  {\Best{7.56}}
  {\Best{9.03}}
  {\Best{7.29}}
& \CSFRTuple
  {\Second{7.13}}
  {\Second{11.29}}
  {\Second{12.99}}
  {\Second{10.47}}
\\

& CAGE
& \Second{45.01}
& \CSFRTuple
  {\Best{7.67}}
  {\Best{13.25}}
  {\Best{17.45}}
  {\Best{12.79}}
& \CSFRTuple
  {\Second{3.82}}
  {\Second{5.61}}
  {\Second{8.19}}
  {\Second{5.87}}
& \CSFRTuple
  {\Second{3.48}}
  {\Second{5.33}}
  {\Second{8.49}}
  {\Second{5.77}}
& \CSFRTuple
  {\Best{9.68}}
  {\Best{13.31}}
  {\Best{16.35}}
  {\Best{13.11}}
\\

& CAA$^{\dagger}$
& 9.85
& \CSFRTuple{4.35}{7.50}{8.78}{6.88}
& \CSFRTuple{1.90}{4.14}{6.66}{4.23}
& \CSFRTuple{0.77}{2.21}{2.65}{1.88}
& \CSFRTuple{7.26}{20.03}{28.32}{18.54}
\\

& \textbf{CIRA}
& \Best{11.68}
& \CSFRTuple
  {4.87}
  {7.90}
  {\Second{11.65}}
  {8.14}
& \CSFRTuple
  {3.04}
  {4.01}
  {\Best{8.42}}
  {5.16}
& \CSFRTuple{2.83}{3.26}{4.75}{3.61}
& \CSFRTuple{3.82}{8.60}{12.26}{8.23}
\\

\Xhline{1.1pt}

% InternVL: TextVQA
\multirow{4}{*}{\textbf{TextVQA}}
& VEAttack
& 56.92
& \CSFRTuple{7.50}{12.77}{14.81}{11.69}
& \CSFRTuple
  {3.83}
  {7.17}
  {\Second{11.67}}
  {7.56}
& \CSFRTuple
  {\Second{7.52}}
  {\Second{10.50}}
  {11.86}
  {9.96}
& \CSFRTuple
  {\Second{5.24}}
  {7.37}
  {12.68}
  {8.43}
\\

& CAGE
& \Second{55.24}
& \CSFRTuple
  {\Second{11.00}}
  {\Second{16.23}}
  {\Second{16.24}}
  {\Second{14.49}}
& \CSFRTuple
  {\Second{6.20}}
  {\Second{10.99}}
  {11.41}
  {\Second{9.53}}
& \CSFRTuple
  {6.64}
  {10.37}
  {\Second{13.50}}
  {\Second{10.17}}
& \CSFRTuple
  {4.94}
  {\Second{9.89}}
  {\Second{14.83}}
  {\Second{9.89}}
\\

& CAA$^{\dagger}$
& 5.03
& \CSFRTuple{10.58}{22.73}{23.58}{18.96}
& \CSFRTuple{13.64}{15.66}{23.68}{17.66}
& \CSFRTuple{5.25}{9.51}{10.53}{8.43}
& \CSFRTuple{10.18}{33.63}{45.69}{29.84}
\\

& \textbf{CIRA}
& \Best{12.03}
& \CSFRTuple
  {\Best{33.17}}
  {\Best{36.80}}
  {\Best{36.93}}
  {\Best{35.63}}
& \CSFRTuple
  {\Best{10.55}}
  {\Best{17.00}}
  {\Best{21.58}}
  {\Best{16.38}}
& \CSFRTuple
  {\Best{17.35}}
  {\Best{21.20}}
  {\Best{15.79}}
  {\Best{18.11}}
& \CSFRTuple
  {\Best{10.03}}
  {\Best{25.36}}
  {\Best{38.28}}
  {\Best{24.56}}
\\

\Xhline{1.1pt}

% InternVL: MME
\multirow{4}{*}{\textbf{MME}}
& VEAttack
& 40.63
& \CSFRTuple{2.57}{4.44}{4.91}{3.98}
& \CSFRTuple{2.27}{4.05}{4.44}{3.58}
& \CSFRTuple{3.28}{3.93}{5.71}{4.31}
& \CSFRTuple{1.72}{3.27}{5.35}{3.45}
\\

& CAGE
& \Second{37.25}
& \CSFRTuple
  {\Best{7.13}}
  {\Second{9.14}}
  {\Second{9.16}}
  {\Second{8.48}}
& \CSFRTuple
  {\Best{4.07}}
  {\Second{4.93}}
  {\Second{6.32}}
  {\Second{5.10}}
& \CSFRTuple
  {\Second{4.38}}
  {\Best{5.73}}
  {\Best{7.82}}
  {\Second{5.98}}
& \CSFRTuple
  {\Best{4.70}}
  {\Best{7.02}}
  {\Second{8.43}}
  {\Second{6.72}}
\\

& CAA$^{\dagger}$
& 5.53
& \CSFRTuple{2.92}{3.58}{8.11}{4.87}
& \CSFRTuple{2.41}{6.19}{8.31}{5.64}
& \CSFRTuple{2.81}{2.81}{5.04}{3.55}
& \CSFRTuple{3.44}{13.68}{18.88}{12.00}
\\

& \textbf{CIRA}
& \Best{5.76}
& \CSFRTuple
  {\Second{6.88}}
  {\Best{10.63}}
  {\Best{12.50}}
  {\Best{10.00}}
& \CSFRTuple
  {\Second{3.98}}
  {\Best{7.45}}
  {\Best{8.72}}
  {\Best{6.72}}
& \CSFRTuple
  {\Best{5.37}}
  {\Second{5.61}}
  {\Second{7.17}}
  {\Best{6.05}}
& \CSFRTuple
  {\Second{4.01}}
  {\Second{6.54}}
  {\Best{11.38}}
  {\Best{7.31}}
\\

\Xhline{2pt}

\end{tabular}
\end{adjustbox}

\end{table*}

%% file: appendix/defense.tex
\section{Selection Stabilization Defense}
\label{sec:appendix_defense}

Translation-Consensus Selection (TCS) stabilizes priority rankings by
aggregating aligned scores across spatially translated views.

\subsection{Translation-Consensus Selection}
\label{sec:appendix_tcs}

For pixel displacement $d$, define
\begin{equation}
    \mathcal{V}
    =\{T_{0,0},T_{d,0},T_{0,d},T_{d,d}\}.
    \label{eq:tcs_views}
\end{equation}
We set $d=7$ pixels and construct each view by reflection-padding and cropping
to the original size. The four views form one batched vision-encoder input.

\paragraph{Score alignment and consensus.}

For view $v\in\mathcal{V}$, let $\mathbf{s}^{(v)}\in\mathbb{R}^{N}$ be its
encoder-side priority-score vector. Operator $\mathcal{A}_{v}$
inverse-aligns the score grid to $T_{0,0}$ by bilinear sampling with reflection
padding. With patch size $p=14$, the offset is $d/p=0.5$ patch:
\begin{equation}
    \widetilde{\mathbf{s}}^{(v)}
    =\mathcal{A}_{v}\!\left(\mathbf{s}^{(v)}\right).
    \label{eq:tcs_alignment}
\end{equation}
\Eqref{eq:tcs_main} converts the aligned scores to descending rank
quantiles, so each view contributes a priority ranking rather than a score scale.

\paragraph{Selection interface.}
At compression budget $K$, TCS replaces the original priority ranking with the
cross-view consensus ranking. The unshifted view supplies token features and
the compressor's key-similarity metric, while translated views contribute
aligned priority scores. Token counts, aggregation, and language-model input
length remain unchanged.

\paragraph{Matched evaluation and cost.}
We evaluate None and TCS on matched adversarial images, questions, references,
eligibility sets, and compression budgets. Full-token inference is unchanged,
so Full ASR is shared within each matched pair in
Table~\ref{tab:tcs_defense}. The four views are processed by the vision encoder
in one batch, with no additional language-model inference.

\subsection{Cross-View Support Mechanism}
\label{sec:appendix_tcs_mechanism}

Cross-view support characterizes the contrast between stable clean evidence
and view-specific adversarial replacements. At each compression budget $K$, the canonical clean Top-$K$ set serves as
the reference, while CIRA replacements are tokens that enter the canonical
adversarial Top-$K$ set from outside this reference. A candidate's
\emph{view support} is the number of aligned views in which it remains within
the Top-$K$ set.

\begin{figure}[htbp]
    \centering
    \includegraphics[width=\textwidth]{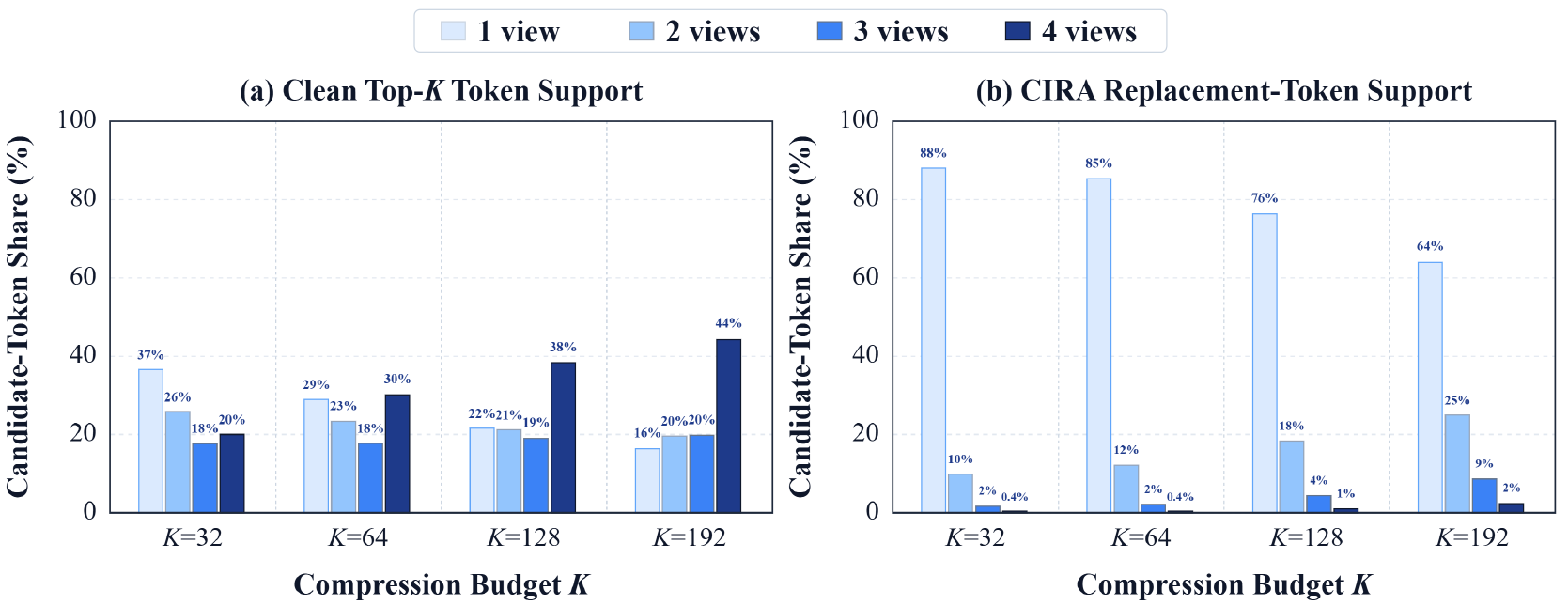}
    \caption{Cross-view support distributions of clean Top-$K$ tokens and CIRA
    replacement tokens across compression budgets.}
    \label{fig:tcs_view_support}
\end{figure}

Figure~\ref{fig:tcs_view_support} shows that CIRA replacements are
predominantly view-specific. As $K$ increases from 32 to 192,
the one-view share decreases from $88.0\%$ to $63.9\%$, while fewer than
$2.5\%$ are supported by all four views. Clean Top-$K$ tokens show the opposite
pattern: their four-view share increases from $20.0\%$ to $44.2\%$. Averaging
aligned rank quantiles therefore downweights isolated replacement spikes while
favoring evidence supported across translations.

\subsection{Adaptive Evaluation}
\label{sec:appendix_adaptive}

Standard CIRA is optimized on the unshifted view, with TCS applied only at
evaluation. Adaptive CIRA instead optimizes
\eqref{eq:adaptive_multiview} over all four public transformations, using the
view-specific encoder objectives as differentiable surrogates for rank
conversion and Top-$K$ selection. For each view, we cache clean features and
ranks and compute GSH and HEP with view-specific hidden-evidence weights over
$[K_{\min},K_{\max}]$. The averaged gradient updates one shared perturbation
using the original $\epsilon=4/255$, step size $1/255$, 100 steps, and
$\lambda=0.8$. Adaptive CIRA retains the same target-encoder access as standard
CIRA.

\paragraph{Selection and rank response.}
For each condition, clean Top-$K$ retention measures the fraction of tokens in
the clean Top-$K$ set that remain in the adversarial Top-$K$ set under the
corresponding ranking rule. Signed normalized rank change is
$(r_i^{a}-r_i^{c})/(N-1)$, where $r_i^{c}$ and $r_i^{a}$ denote clean and
adversarial ranks under the same ranking rule, and positive values indicate
demotion. Retention is averaged per image.  

\begin{figure}[htbp]
    \centering
    \includegraphics[width=\textwidth]{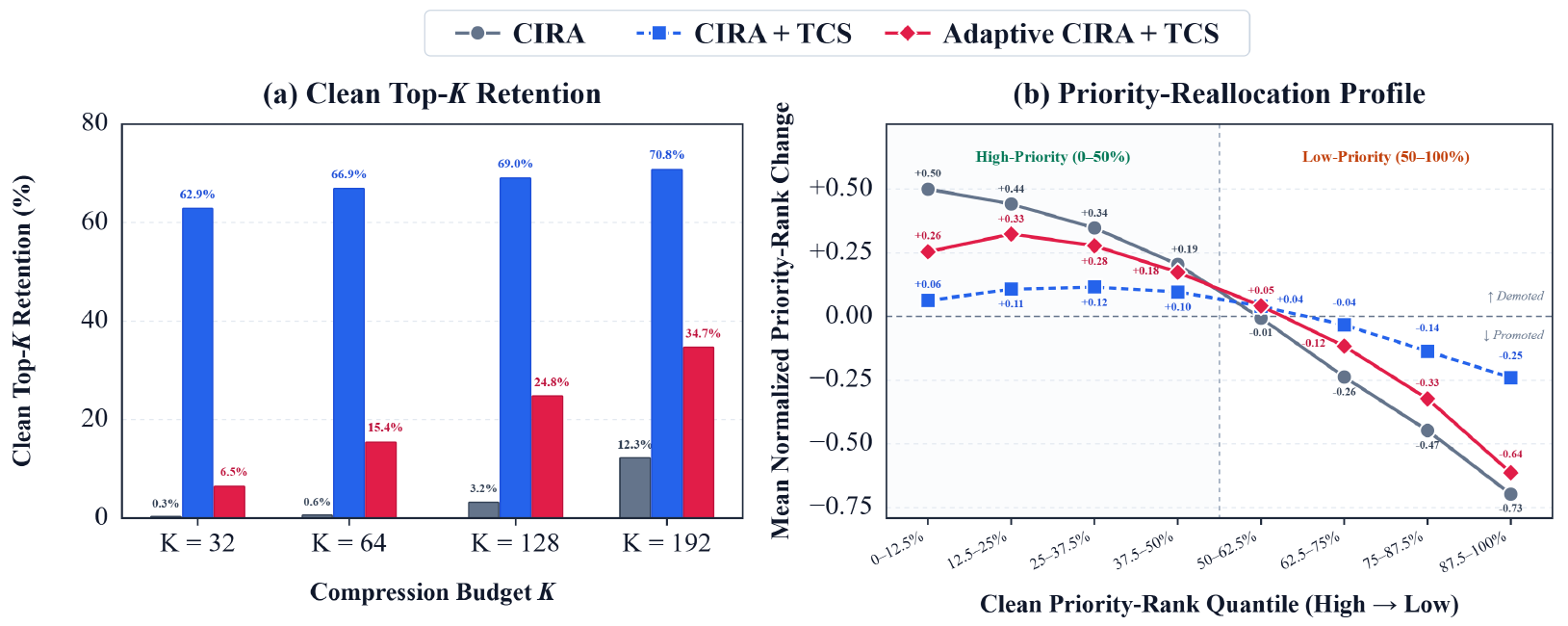}
    \caption{Clean Top-$K$ retention and priority-reallocation profiles under
    CIRA, CIRA + TCS, and Adaptive CIRA + TCS.}
    \label{fig:tcs_selection_mechanism}
\end{figure}

Across the four compression budgets,
Figure~\ref{fig:tcs_selection_mechanism}(a) shows that TCS
raises clean Top-$K$ retention under CIRA from $0.3$--$12.3\%$ to
$62.9$--$70.8\%$. Adaptive CIRA reduces this retention under TCS to
$6.5$--$34.7\%$. Figure~\ref{fig:tcs_selection_mechanism}(b) shows the corresponding
priority reallocation: TCS attenuates both the demotion of clean high-priority
tokens and the promotion of initially low-priority tokens, whereas Adaptive
CIRA restores much of this signed reallocation.

Together, the cross-view support patterns and the selection responses under TCS
show that TCS suppresses view-fragile priority reallocation.
Corresponding task-utility results are reported in
\AppendixRef{sec:appendix_task_utility}.

%% file: appendix/utility.tex
\section{Complementary Task-Utility Results}
\label{sec:appendix_task_utility}

CSFR is the primary clean-conditioned metric for compression-specific failure.
We complement it with accuracy-based results that characterize clean utility
under compression, post-attack performance, and task-utility recovery under TCS.

\subsection{Clean Utility under Compression}

\input{tables/clean_utility}

Table~\ref{tab:clean_utility} shows that high-budget settings preserve
most POPE and MME accuracy across model families, while TextVQA generally
exhibits larger losses under compression.
Related analyses also find task-dependent visual-token requirements,
with OCR tasks relying on visual information deeper into the decoder
\citep{wang2026token}.
These values provide the clean reference for the post-attack comparisons below.

\subsection{Post-Attack Task Utility}

Unlike CSFR, adversarial accuracy is computed over the complete evaluation set
and therefore reflects both pre-existing compression errors and attack-induced
failures. It provides a complementary view of overall task degradation under CIRA.

\input{tables/attacked_utility}

Table~\ref{tab:attacked_utility} shows that CIRA reduces budget-averaged
compressed accuracy by $3.3$--$20.8$ pp across the evaluated
model--dataset--compressor settings. After averaging compressors within each
model--dataset pair and weighting the nine pairs equally, the mean compressed
accuracy drop is $9.6$ pp, compared with $5.4$ pp under full-token inference.
This aggregate view complements the paired CSFR analysis by quantifying
end-task degradation under compressed inference.

\input{tables/tcs_utility}

\subsection{Task-Utility Recovery under TCS}

Table~\ref{tab:tcs_utility} reports matched task accuracy without and with TCS.
Each comparison fixes the input image, question, and compression budget, with
TCS applied only to compressed inference.

Across the three benchmarks, TCS changes clean Avg.\ ACC by at most $0.52$ pp
in magnitude, while recovering $6.77$--$13.20$ pp under CIRA, with larger gains
at tighter budgets. Under Adaptive CIRA, TCS recovers less utility, consistent
with the attack partially restoring the priority reallocation suppressed by
TCS.

%% file: tables/clean_utility.tex
\begin{table*}[htbp]
\centering
\scriptsize

\caption{Clean full-token and compressed accuracy (\%) across model families,
datasets, compressors, and family-specific compression budgets. The final value
in each compressed cell reports the average across the listed budgets.}
\label{tab:clean_utility}

\vspace{1pt}
\renewcommand{\arraystretch}{1.13}
\setlength{\tabcolsep}{4.2pt}

\begin{adjustbox}{max width=\textwidth}
\begin{tabular}{
l
!{\color{black}\vrule width 0.45pt}
c
!{\color{black}\vrule width 0.45pt}
cccc
}
\Xhline{2pt}
\rowcolor{headercolor}
\textbf{Dataset}
& \textbf{Full ACC}
& \textbf{VisionZip}
& \textbf{VisPruner}
& \textbf{PruMerge}
& \textbf{FastV}
\\
\Xhline{2pt}

\multicolumn{6}{>{\columncolor{gray!20}}c}{
\textbf{LLaVA-v1.5-7B}
\quad ($K=192/128/64/32/\mathrm{Avg.}$)
}
\\
\hline
POPE
& 84.6
& 84.3/83.5/79.6/73.0/80.1
& 84.1/82.9/80.5/75.3/80.7
& 75.6/74.1/72.0/70.1/73.0
& 81.1/80.0/74.7/68.0/76.0
\\
TextVQA
& 60.3
& 54.3/51.6/49.5/45.1/50.1
& 58.4/58.0/56.8/51.9/56.3
& 51.9/51.6/51.7/49.4/51.2
& 51.6/49.4/44.7/37.8/45.9
\\
MME
& 79.2
& 77.2/76.3/73.5/69.0/74.0
& 76.9/76.5/74.0/71.7/74.8
& 72.9/70.9/71.1/70.8/71.4
& 76.1/73.9/70.8/67.4/72.1
\\
\Xhline{1.2pt}

\multicolumn{6}{>{\columncolor{gray!20}}c}{
\textbf{Qwen3-VL-8B-Instruct}
\quad ($K=96/64/32/\mathrm{Avg.}$)
}
\\
\hline
POPE
& 86.3
& 86.1/84.1/80.7/83.6
& 86.6/85.1/82.3/84.7
& 86.4/86.1/83.8/85.4
& 85.1/82.4/74.4/80.6
\\
TextVQA
& 88.6
& 46.7/41.1/34.1/40.6
& 45.8/42.6/41.5/43.3
& 33.9/31.8/31.2/32.3
& 52.6/40.6/28.1/40.4
\\
MME
& 90.0
& 88.2/88.6/83.4/86.7
& 88.4/88.1/83.5/86.7
& 86.7/86.7/83.3/85.6
& 88.3/85.7/79.4/84.5
\\
\Xhline{1.2pt}

\multicolumn{6}{>{\columncolor{gray!20}}c}{
\textbf{InternVL3.5-8B}
\quad ($K=128/64/32/\mathrm{Avg.}$)
}
\\
\hline
POPE
& 82.2
& 81.4/80.8/77.1/79.8
& 80.4/80.5/78.3/79.7
& 79.0/79.0/78.8/78.9
& 81.0/79.5/74.9/78.5
\\
TextVQA
& 71.5
& 64.7/48.6/37.5/50.3
& 57.7/47.0/40.4/48.4
& 46.8/37.7/32.8/39.1
& 68.5/57.6/44.3/56.8
\\
MME
& 88.6
& 87.5/83.8/78.2/83.2
& 84.8/82.2/77.0/81.3
& 84.0/80.8/78.5/81.1
& 88.2/85.2/77.8/83.7
\\
\Xhline{2pt}
\end{tabular}
\end{adjustbox}

\vspace{-3pt}
\end{table*}

%% file: tables/attacked_utility.tex
\begin{table*}[htbp]
\centering
\scriptsize

\caption{Full-token clean-to-adversarial accuracy and compressed adversarial
accuracy (\%) under CIRA across model families, datasets, compressors, and
family-specific budgets. The final value in each compressed cell reports the
budget average.}

\label{tab:attacked_utility}

\vspace{1pt}
\renewcommand{\arraystretch}{1.13}
\setlength{\tabcolsep}{5.0pt}

\begin{adjustbox}{max width=\textwidth}
\begin{tabular}{
l
!{\color{black}\vrule width 0.45pt}
c
!{\color{black}\vrule width 0.45pt}
cccc
}
\Xhline{2pt}
\rowcolor{headercolor}
\textbf{Dataset}
& \textbf{Full ACC}
& \textbf{VisionZip}
& \textbf{VisPruner}
& \textbf{PruMerge}
& \textbf{FastV}
\\
\Xhline{2pt}

\multicolumn{6}{>{\columncolor{gray!20}}c}{
\textbf{LLaVA-v1.5-7B}
\quad (compressed Adv.\ ACC: $K=192/128/64/32/\mathrm{Avg.}$)
}
\\
\hline
POPE
& $84.6\rightarrow82.6$
& $78.4/74.0/64.6/58.0/68.8$
& $80.0/77.7/71.7/64.1/73.4$
& $63.3/55.9/51.2/50.3/55.2$
& $76.8/71.4/61.0/53.4/65.7$
\\
TextVQA
& $60.3\rightarrow57.1$
& $46.7/40.0/32.2/26.1/36.3$
& $51.3/49.5/44.6/38.6/46.0$
& $36.9/32.2/27.4/25.0/30.4$
& $45.5/41.9/33.5/25.7/36.7$
\\
MME
& $79.2\rightarrow77.2$
& $73.2/70.7/65.5/59.5/67.2$
& $73.2/70.9/64.1/60.5/67.2$
& $62.2/58.9/55.4/51.5/57.0$
& $73.2/71.6/63.7/58.8/66.8$
\\
\Xhline{1.2pt}

\multicolumn{6}{>{\columncolor{gray!20}}c}{
\textbf{Qwen3-VL-8B-Instruct}
\quad (compressed Adv.\ ACC: $K=96/64/32/\mathrm{Avg.}$)
}
\\
\hline
POPE
& $86.3\rightarrow81.3$
& $75.9/74.2/71.2/73.8$
& $78.1/76.7/73.5/76.1$
& $79.8/79.8/76.0/78.5$
& $76.0/73.9/67.7/72.5$
\\
TextVQA
& $88.6\rightarrow75.7$
& $21.1/20.5/21.4/21.0$
& $31.9/31.9/31.0/31.6$
& $22.2/22.4/22.0/22.2$
& $30.7/25.6/20.6/25.6$
\\
MME
& $90.0\rightarrow81.5$
& $74.9/76.1/69.8/73.6$
& $73.9/76.5/72.5/74.3$
& $74.3/77.1/74.6/75.3$
& $79.9/77.5/70.9/76.1$
\\
\Xhline{1.2pt}

\multicolumn{6}{>{\columncolor{gray!20}}c}{
\textbf{InternVL3.5-8B}
\quad (compressed Adv.\ ACC: $K=128/64/32/\mathrm{Avg.}$)
}
\\
\hline
POPE
& $82.2\rightarrow76.1$
& $75.5/75.4/71.8/74.2$
& $75.8/75.9/73.4/75.0$
& $75.7/75.0/75.4/75.4$
& $77.3/73.1/70.2/73.5$
\\
TextVQA
& $71.5\rightarrow65.9$
& $42.3/32.3/27.1/33.9$
& $53.9/43.6/36.1/44.5$
& $39.5/31.8/29.5/33.6$
& $58.4/42.7/29.6/43.6$
\\
MME
& $88.6\rightarrow85.3$
& $81.4/75.7/71.5/76.2$
& $81.3/77.8/74.9/78.0$
& $80.2/78.3/74.7/77.7$
& $83.8/79.8/72.9/78.8$
\\
\Xhline{2pt}
\end{tabular}
\end{adjustbox}

\vspace{-3pt}
\end{table*}

%% file: tables/tcs_utility.tex
\begin{table}[htbp]
\centering
\scriptsize

\caption{Matched compressed accuracy (\%) without and with TCS on
LLaVA-v1.5-7B with VisionZip under clean, CIRA, and Adaptive CIRA conditions
across datasets and compression budgets. Entries show
$\mathrm{None}\!\rightarrow\!\mathrm{TCS}$, with annotations reporting the
corresponding change.}
\label{tab:tcs_utility}

\vspace{1pt}
\renewcommand{\arraystretch}{1.13}
\setlength{\tabcolsep}{5.0pt}

\begin{adjustbox}{max width=\textwidth}
\begin{tabular}{
l
!{\color{black}\vrule width 0.45pt}
cccc
!{\color{black}\vrule width 0.45pt}
c
}
\Xhline{2pt}
\rowcolor{headercolor}
\textbf{Evaluation}
& $K=192$
& $K=128$
& $K=64$
& $K=32$
& \textbf{Avg.}
\\
\Xhline{2pt}

\multicolumn{6}{>{\columncolor{gray!20}}c}{\textbf{POPE}}
\\
Clean
& $84.30\!\rightarrow\!83.10$\,\DefenseAccDown{1.20}
& $83.50\!\rightarrow\!82.50$\,\DefenseAccDown{1.00}
& $79.60\!\rightarrow\!78.40$\,\DefenseAccDown{1.20}
& $73.00\!\rightarrow\!74.30$\,\DefenseAccUp{1.30}
& $80.10\!\rightarrow\!79.58$\,\DefenseAccDown{0.52}
\\
CIRA
& $78.40\!\rightarrow\!81.30$\,\DefenseAccUp{2.90}
& $74.00\!\rightarrow\!80.90$\,\DefenseAccUp{6.90}
& $64.60\!\rightarrow\!78.30$\,\DefenseAccUp{13.70}
& $58.00\!\rightarrow\!75.40$\,\DefenseAccUp{17.40}
& $68.75\!\rightarrow\!78.98$\,\DefenseAccUp{10.23}
\\
Adaptive CIRA
& $78.80\!\rightarrow\!78.60$\,\DefenseAccDown{0.20}
& $76.40\!\rightarrow\!76.00$\,\DefenseAccDown{0.40}
& $66.80\!\rightarrow\!71.60$\,\DefenseAccUp{4.80}
& $56.30\!\rightarrow\!62.20$\,\DefenseAccUp{5.90}
& $69.58\!\rightarrow\!72.10$\,\DefenseAccUp{2.52}
\\
\Xhline{0.45pt}

\multicolumn{6}{>{\columncolor{gray!20}}c}{\textbf{TextVQA}}
\\
Clean
& $54.30\!\rightarrow\!53.20$\,\DefenseAccDown{1.10}
& $51.60\!\rightarrow\!51.30$\,\DefenseAccDown{0.30}
& $49.50\!\rightarrow\!49.40$\,\DefenseAccDown{0.10}
& $45.10\!\rightarrow\!45.40$\,\DefenseAccUp{0.30}
& $50.13\!\rightarrow\!49.83$\,\DefenseAccDown{0.30}
\\
CIRA
& $46.70\!\rightarrow\!52.40$\,\DefenseAccUp{5.70}
& $40.00\!\rightarrow\!50.90$\,\DefenseAccUp{10.90}
& $32.20\!\rightarrow\!49.00$\,\DefenseAccUp{16.80}
& $26.10\!\rightarrow\!45.50$\,\DefenseAccUp{19.40}
& $36.25\!\rightarrow\!49.45$\,\DefenseAccUp{13.20}
\\
Adaptive CIRA
& $48.30\!\rightarrow\!49.00$\,\DefenseAccUp{0.70}
& $45.10\!\rightarrow\!46.90$\,\DefenseAccUp{1.80}
& $36.00\!\rightarrow\!42.10$\,\DefenseAccUp{6.10}
& $26.70\!\rightarrow\!33.00$\,\DefenseAccUp{6.30}
& $39.03\!\rightarrow\!42.75$\,\DefenseAccUp{3.72}
\\
\Xhline{0.45pt}

\multicolumn{6}{>{\columncolor{gray!20}}c}{\textbf{MME}}
\\
Clean
& $77.20\!\rightarrow\!76.50$\,\DefenseAccDown{0.70}
& $76.30\!\rightarrow\!76.00$\,\DefenseAccDown{0.30}
& $73.50\!\rightarrow\!74.30$\,\DefenseAccUp{0.80}
& $69.00\!\rightarrow\!70.60$\,\DefenseAccUp{1.60}
& $74.00\!\rightarrow\!74.35$\,\DefenseAccUp{0.35}
\\
CIRA
& $73.20\!\rightarrow\!75.00$\,\DefenseAccUp{1.80}
& $70.70\!\rightarrow\!75.80$\,\DefenseAccUp{5.10}
& $65.50\!\rightarrow\!74.00$\,\DefenseAccUp{8.50}
& $59.50\!\rightarrow\!71.20$\,\DefenseAccUp{11.70}
& $67.23\!\rightarrow\!74.00$\,\DefenseAccUp{6.77}
\\
Adaptive CIRA
& $74.50\!\rightarrow\!73.60$\,\DefenseAccDown{0.90}
& $70.90\!\rightarrow\!72.00$\,\DefenseAccUp{1.10}
& $62.90\!\rightarrow\!68.40$\,\DefenseAccUp{5.50}
& $58.50\!\rightarrow\!62.00$\,\DefenseAccUp{3.50}
& $66.70\!\rightarrow\!69.00$\,\DefenseAccUp{2.30}
\\
\Xhline{2pt}
\end{tabular}
\end{adjustbox}

\vspace{-3pt}
\end{table}

%% file: appendix/limitations.tex
\section{Limitations and Future Work}
\label{sec:appendix_limitations}

\paragraph{Temporal and contextual dependencies.}
The present evaluation is restricted to single-image inference.
In video, multi-image, and multi-turn settings, evidence retention
also depends on temporal redundancy and evolving context
\citep{shen2024longvu,li2026vista,wang2026rethinking}.
Run-Length Pruning, for instance, combines temporal redundancy removal
with token distillation \citep{ma2026visual}.
Whether encoder-only priority manipulation remains compression-selective
when evidence is distributed across frames or dialogue turns remains
an open question.

\paragraph{Compression mechanisms beyond token selection.}
Our diagnostics examine retained-set allocation and representation drift,
but do not exhaust the mechanisms underlying compression-induced errors.
Adaptive pruning changes allocation across inputs or layers
\citep{ye2025atp,chen2026variation,li2026transprune};
learned summarization transforms the token representation
\citep{bulat2026compress}; and recoverable routing permits deferred tokens
to re-enter subsequent selection stages \citep{yang2026reroute}.
These mechanisms complicate a description based on a single token ranking.
Moreover, spatial disruption \citep{huang2026n} and positional or
attentional distortion \citep{cho2026improving} are not separately
identified by our retained-set interventions.
Paired CSF evaluation remains applicable, but attributing failures
within these compression mechanisms requires additional diagnostics.

\paragraph{Beyond task correctness.}
CSF is defined through task-level correctness rather than response
safety or targeted attack success.
Accordingly, preserving full-token correctness does not establish
safety preservation, and compressed-path errors do not necessarily
constitute safety-alignment failures.
Targeted manipulation and multimodal jailbreaks
\citep{zhang2025anyattack,qi2024visual,shayegani2024jailbreak}
provide distinct settings for paired evaluation, requiring outcome
criteria tailored to the corresponding security objective.

%% file: appendix/case_studies.tex
\clearpage
\section{Qualitative Case Studies}
\label{sec:case-studies}
% Preserve the previous section label for existing external references.
\label{sec:llava-case-studies}

\begingroup
\setlength{\intextsep}{0pt}
\setlength{\parskip}{0pt}
\captionsetup{font=small,skip=3pt}
% Full-width panels; remove interline glue rather than shrinking the images.
\begin{figure}[H]
\centering

\includegraphics[width=\linewidth]{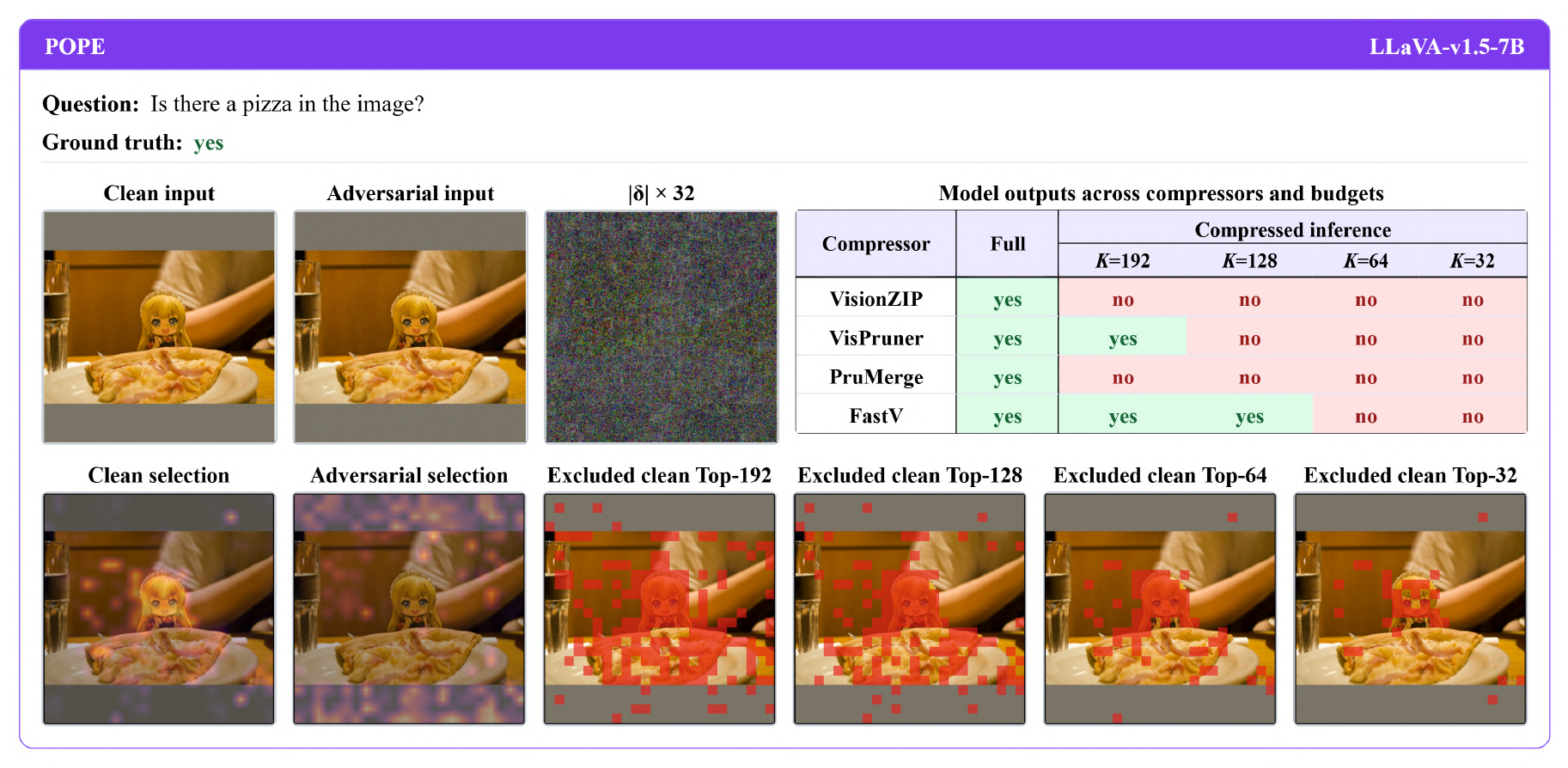}\par
\vspace{1pt}\nointerlineskip
\includegraphics[width=\linewidth]{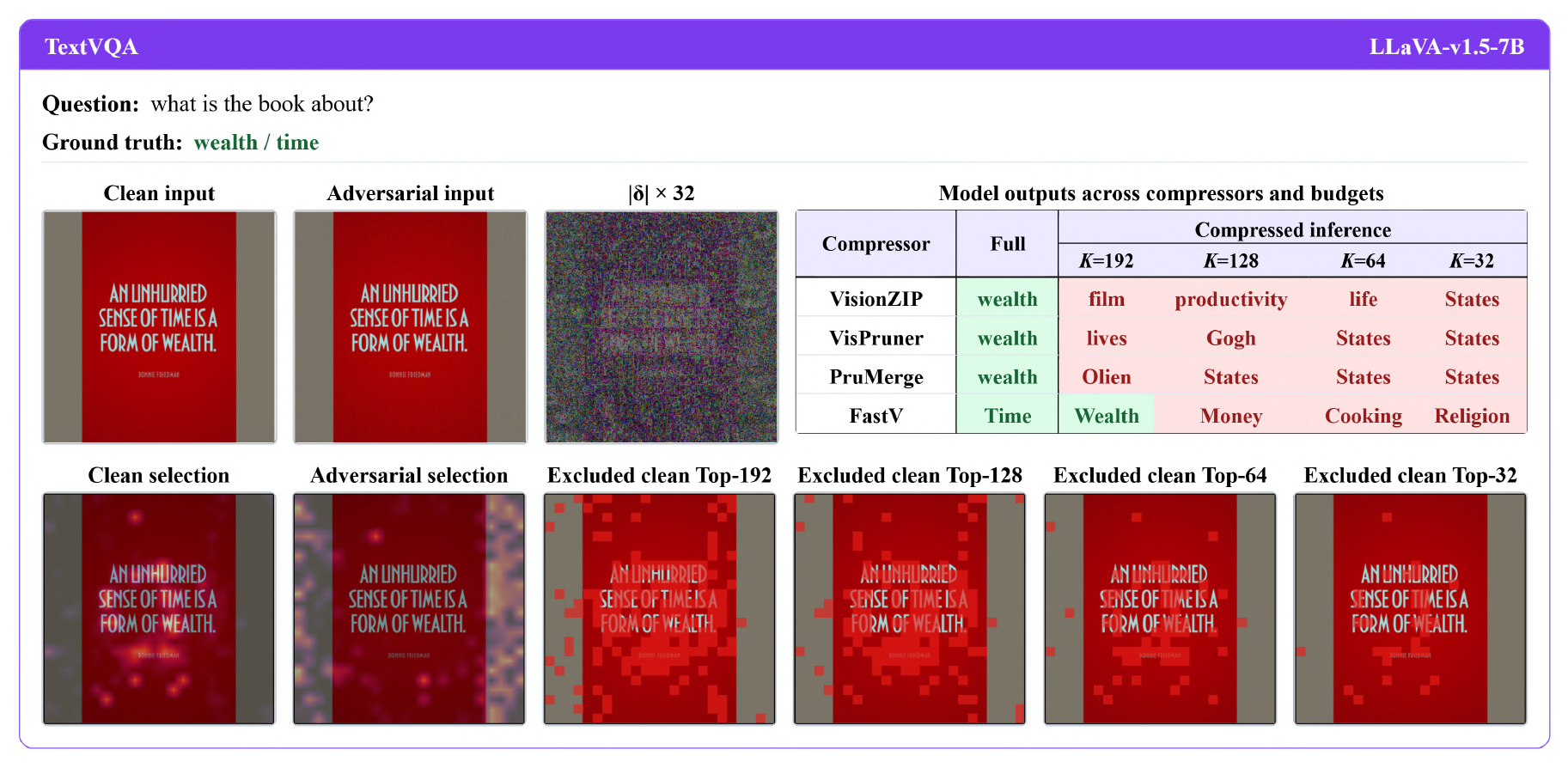}\par
\vspace{1pt}\nointerlineskip
\includegraphics[width=\linewidth]{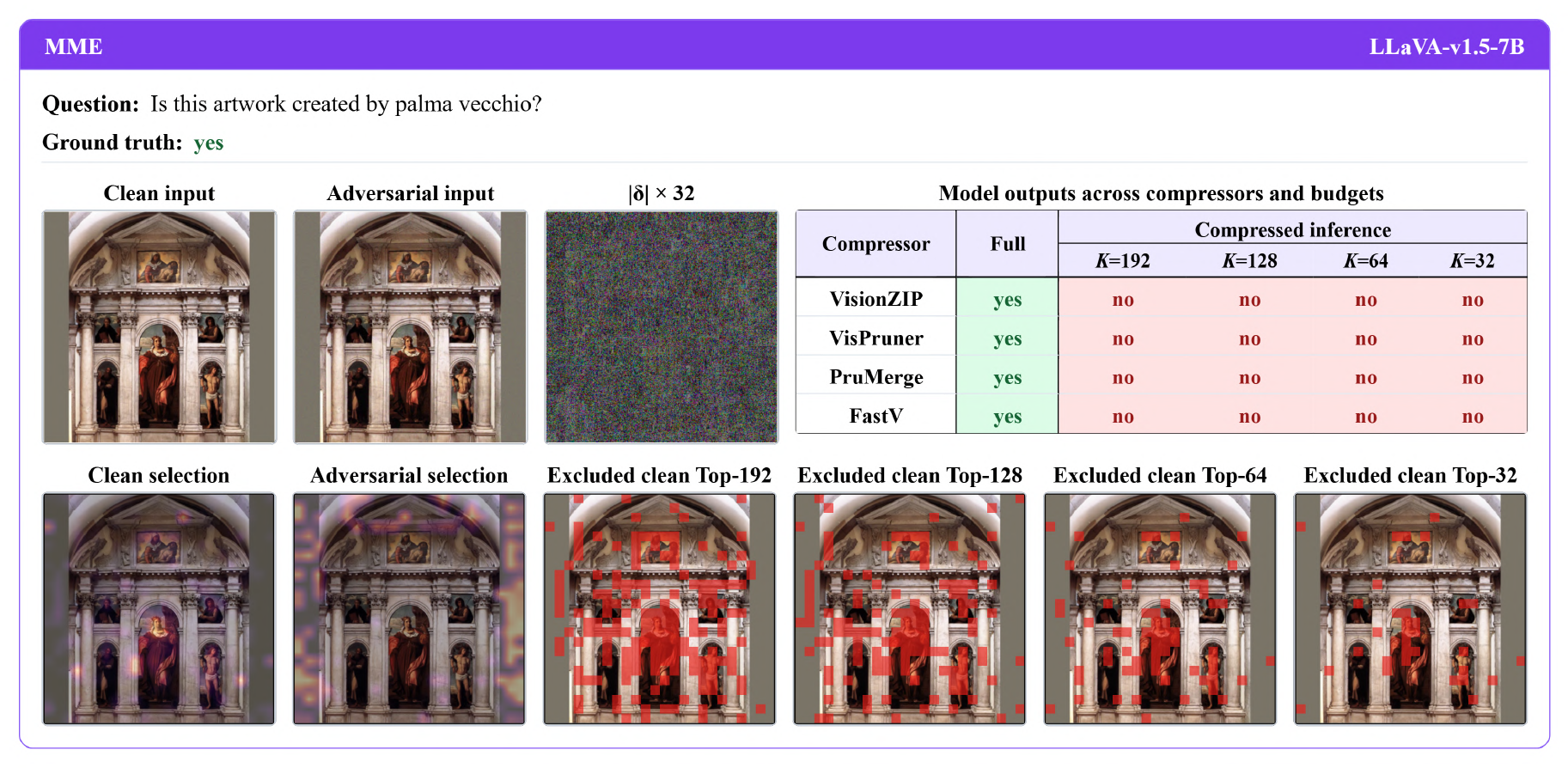}\par

\caption{
Qualitative examples of compression-specific failures on LLaVA-v1.5-7B
across visual-token compressors and retention budgets.
}
\label{fig:llava-cases}
\label{fig:llava-case-page1}

\end{figure}

\clearpage

\begin{figure}[H]
\centering

\includegraphics[width=\linewidth]{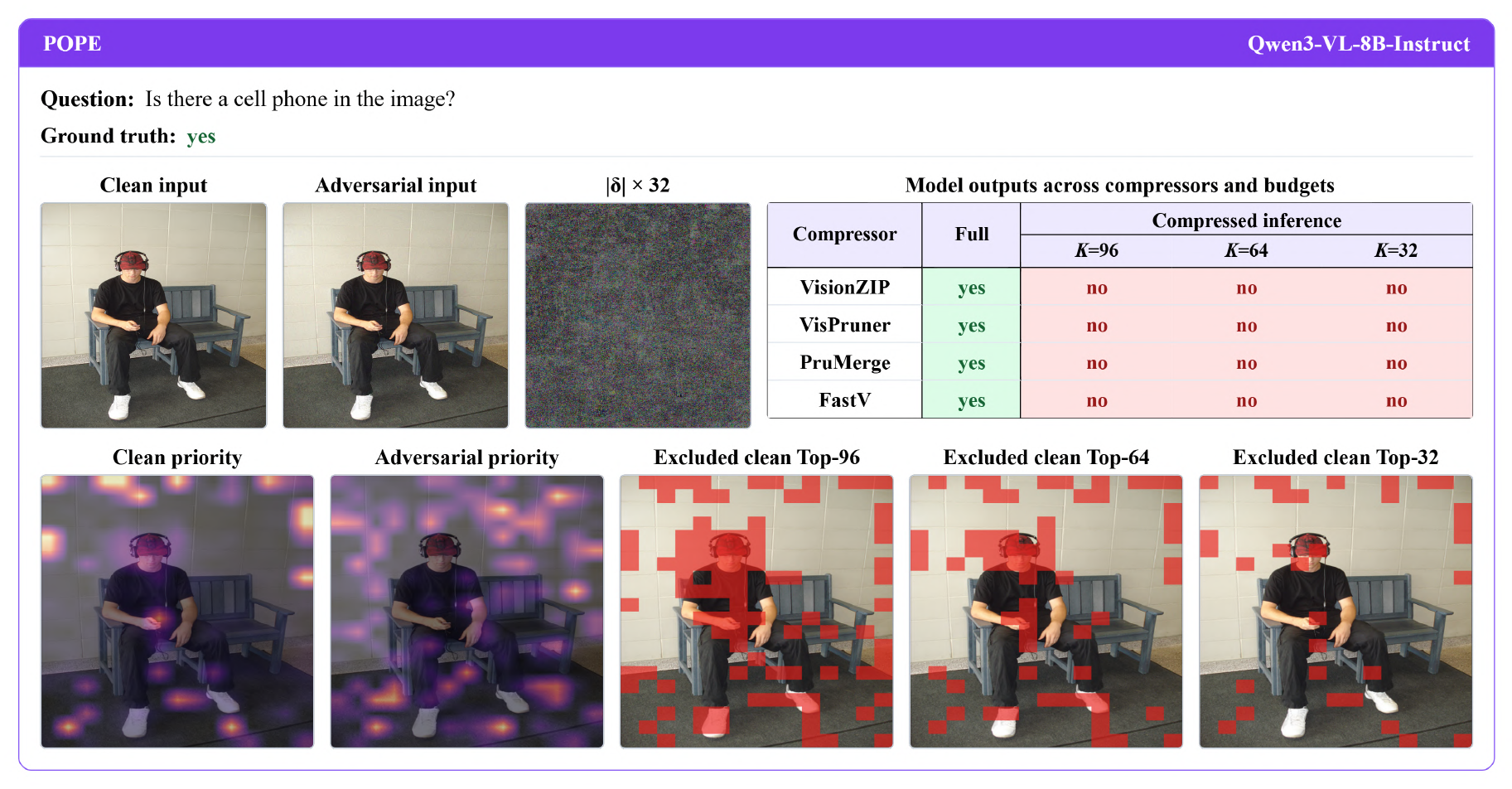}\par
\vspace{1pt}\nointerlineskip
\includegraphics[width=\linewidth]{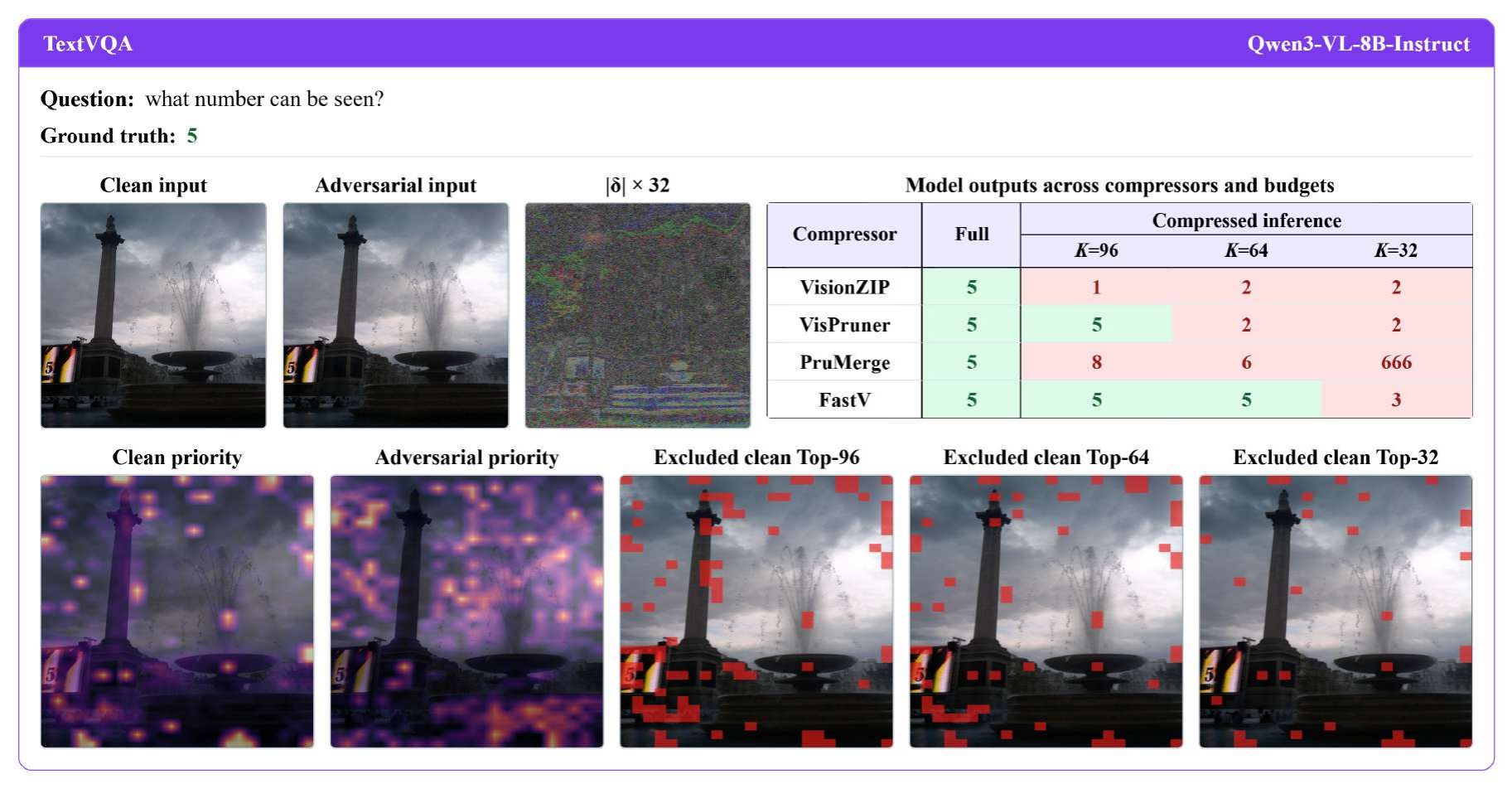}\par
\vspace{1pt}\nointerlineskip
\includegraphics[width=\linewidth]{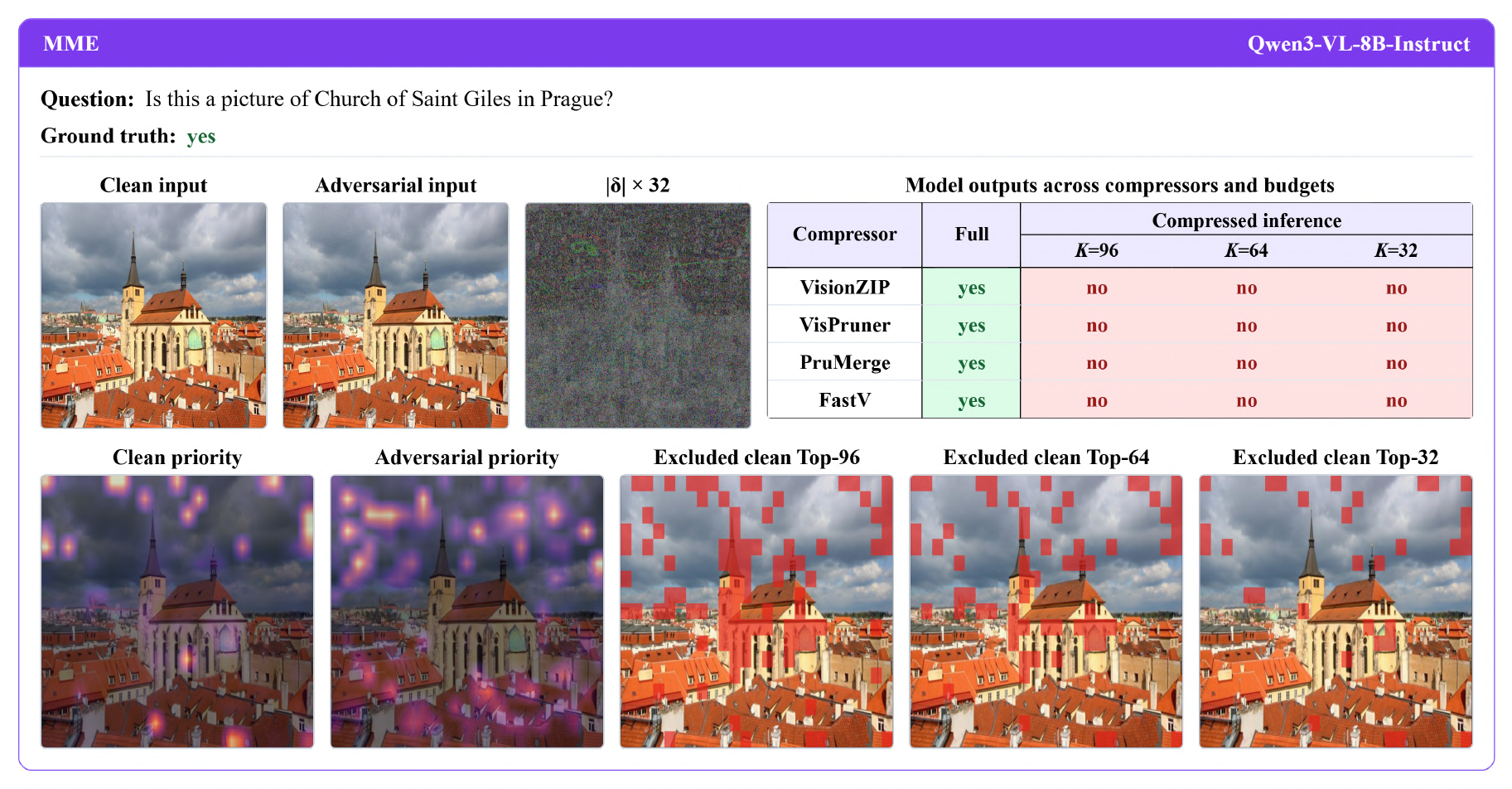}\par

\caption{
Qualitative examples of compression-specific failures on Qwen3-VL-8B-Instruct
across visual-token compressors and retention budgets.
}
\label{fig:qwen-cases}

\end{figure}

\clearpage

\begin{figure}[H]
\centering

\includegraphics[width=\linewidth]{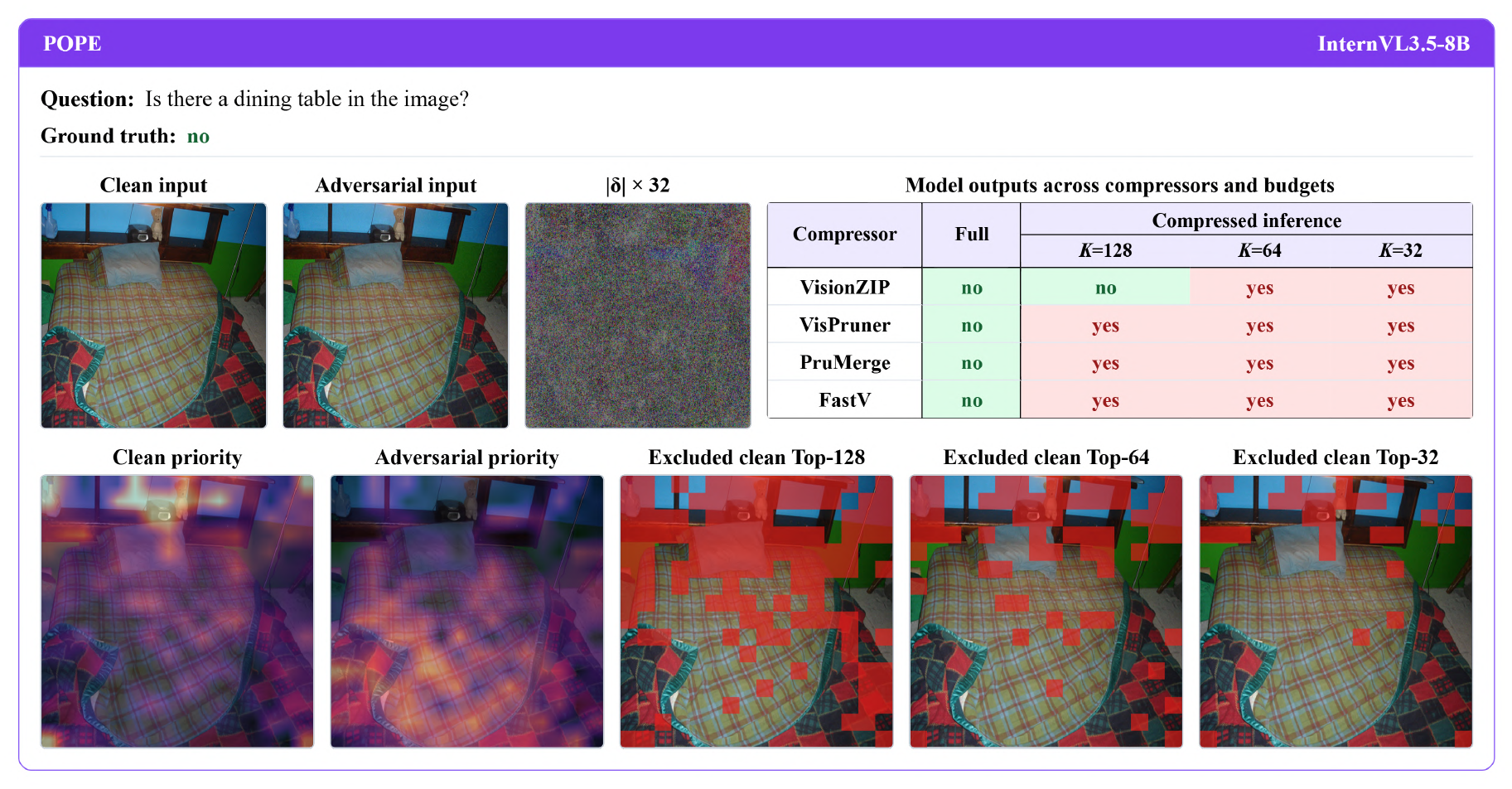}\par
\vspace{1pt}\nointerlineskip
\includegraphics[width=\linewidth]{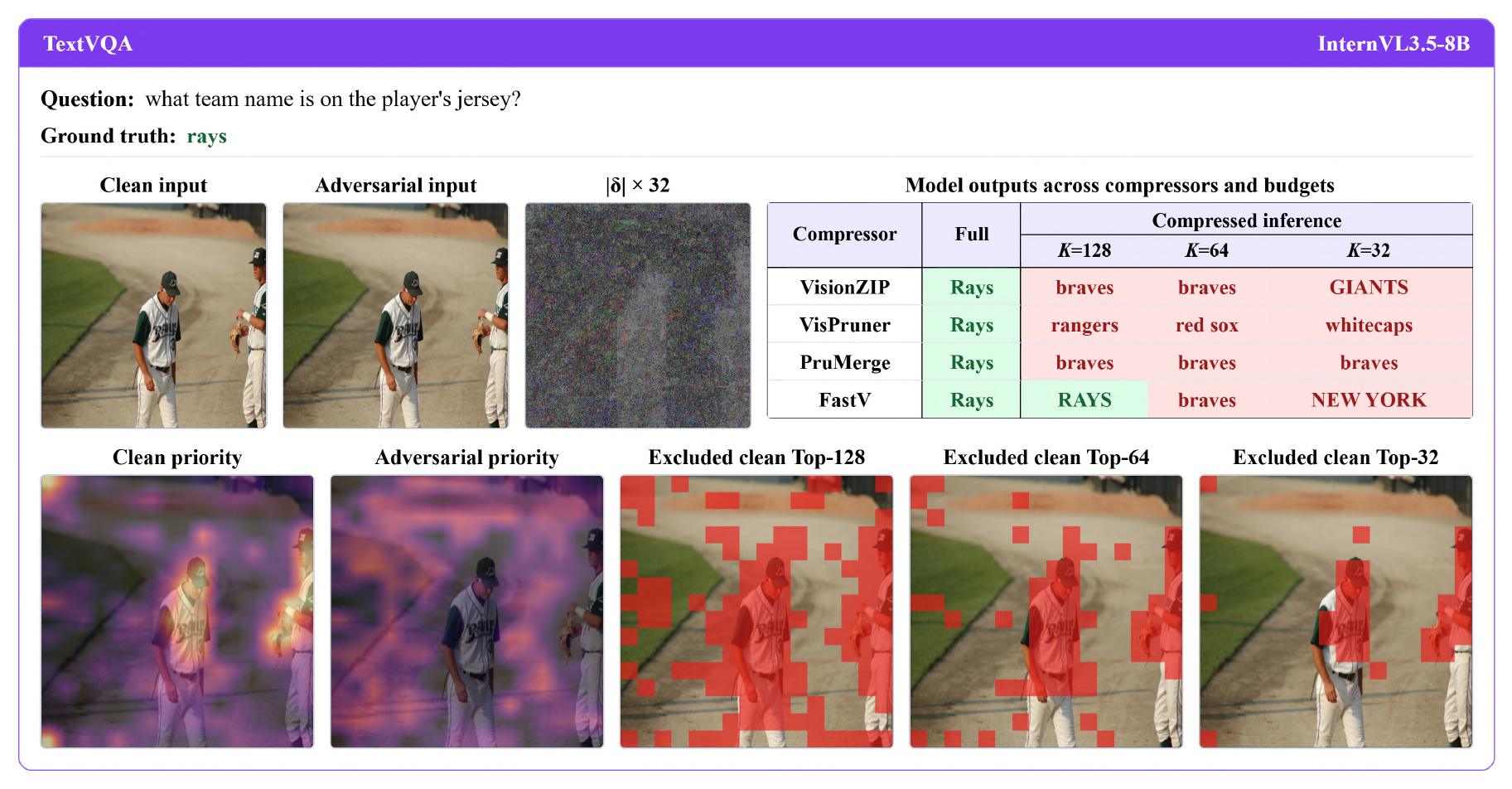}\par
\vspace{1pt}\nointerlineskip
\includegraphics[width=\linewidth]{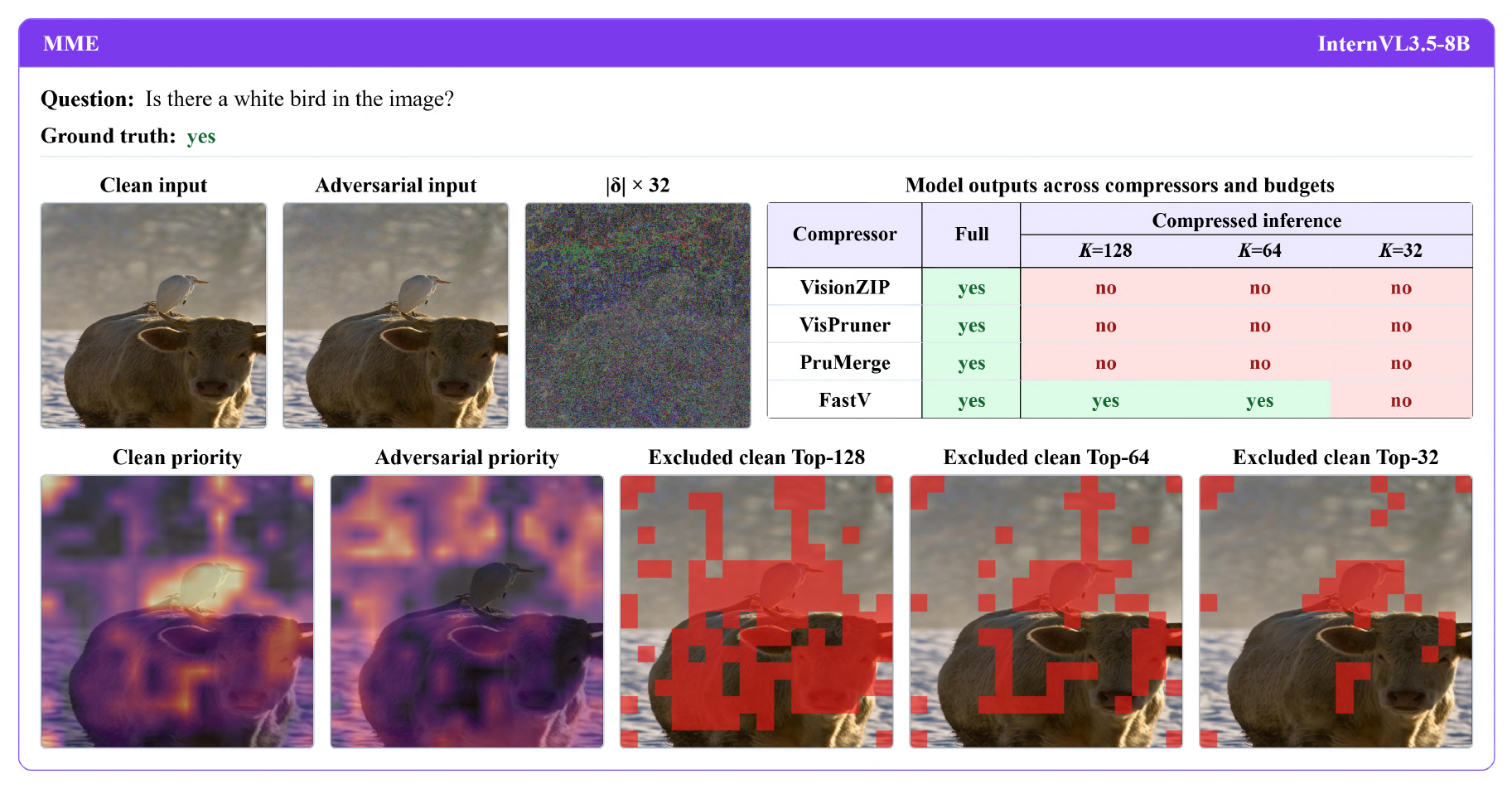}\par

\caption{
Qualitative examples of compression-specific failures on InternVL3.5-8B
across visual-token compressors and retention budgets.
}
\label{fig:internvl-cases}

\end{figure}
\endgroup

\clearpage